\documentclass{article}

\PassOptionsToPackage{numbers, compress}{natbib}

\usepackage[preprint]{neurips_2026}

\usepackage[utf8]{inputenc}
\usepackage[T1]{fontenc}
\usepackage{hyperref}
\usepackage{url}
\usepackage{booktabs}
\usepackage{amsfonts}
\usepackage{nicefrac}
\usepackage{microtype}
\usepackage{xcolor}
\usepackage{graphicx}
\usepackage{tcolorbox}
\usepackage{colortbl}
\usepackage{multirow}
\usepackage{amsmath} 
\usepackage{enumitem}
\usepackage{algorithm}
\usepackage{algpseudocode}
\usepackage{subcaption}
\usepackage{amssymb}
\usepackage{pifont}

\title{EO-WM: A Physically Informed World Model for Probabilistic Earth Observation Forecasting}

\author{%
Junwei Luo$^{1,2}$\thanks{Equal contribution.}
\quad
Shuai Yuan$^{1}$\footnotemark[1]
\quad
Zhenya Yang$^{1}$
\quad
Yansheng Li$^{2}$
\quad \\
\textbf{Zhe Liu}$^{1}$
\quad
\textbf{Hengshuang Zhao}$^{1}$\thanks{Corresponding author.\protect\\ Primary contact to Junwei Luo: \texttt{luojw913@hku.hk}.}\\[6pt]
$^{1}$The University of Hong Kong
\quad
$^{2}$Wuhan University
}

\begin{document}

\maketitle

\begin{abstract}
 Earth Observation (EO) forecasting aims to predict future Earth surface dynamics from satellite observations under changing meteorological conditions. In this paper, we view this task as a partially observed, weather-driven world modeling problem, in which weather acts as a conditioning signal, while forecasting remains uncertain due to sparse observations and unobserved land-surface states. However, existing methods do not fully capture this setting: deterministic models collapse uncertainty into a single future prediction, while diffusion-based methods typically treat weather variables as undifferentiated conditioning signals, and existing benchmarks focus mainly on reconstruction accuracy rather than whether forecasts respond correctly to changed weather forcing.
We introduce EO-WM, a video diffusion transformer for multispectral EO forecasting. EO-WM incorporates a physically informed conditioning framework that represents meteorological forcing through a climatological baseline, weather anomalies, and cumulative physical stress signals. Specifically, it separates baseline and anomaly through distinct conditioning pathways, and accumulates anomalous forcing over time to capture sustained heat and drought stress. To evaluate weather-response behavior beyond standard metrics, we introduce two diagnostic benchmarks: an Extreme Summer Benchmark for severity-aware prediction of vegetation degradation under extreme weather, and a Seasonal Matched-Pair Benchmark for testing response fidelity under changed weather forcing. Experiments show that EO-WM reduces the error in predicted Normalized Difference Vegetation Index (NDVI) decline amplitude by a relative 5.63\% and improves directional hit rate by a relative 7.80\%, while remaining competitive on standard pixel-level metrics. The benchmarks and model will be made open-source at \url{https://github.com/Luo-Z13/EO-WM}.
\end{abstract}

\section{Introduction}

Earth Observation (EO) forecasting predicts future satellite observations conditioned on weather information~\cite{requena2021earthnet2021}, and underpins important downstream applications such as extreme-event monitoring~\cite{trenberth2015attribution,pellicer2025explainable}, crop-yield prediction~\cite{van2020crop}, and ecosystem or vegetation forecasting~\cite{tian2019forecasting,benson2024multi}. Existing work~\cite{gao2022earthformer,Shinohara_2025_vitkoop,zhao2025vegediff} has achieved strong pixel reconstruction quality, yet current formulations and evaluations still capture only part of the EO forecasting problem.

We argue that EO forecasting is more naturally viewed as a partially observed, weather-driven world modeling problem. World models aim to learn predictive state dynamics from observations together with actions or other conditioning inputs~\cite{ha2018world,hafner2019learning,hafner2019dream}, and have become a central paradigm in domains such as game-based visual control environments~\cite{alonso2024diffusion} and autonomous driving~\cite{min2024driveworld,yang2025geniedrive}. EO forecasting likewise requires learning a dynamical process driven by exogenous forcings, except that the conditioning signal is observed weather rather than a controllable agent action.

\begin{figure}[!t]
    \centering
    \includegraphics[width=0.99\textwidth]{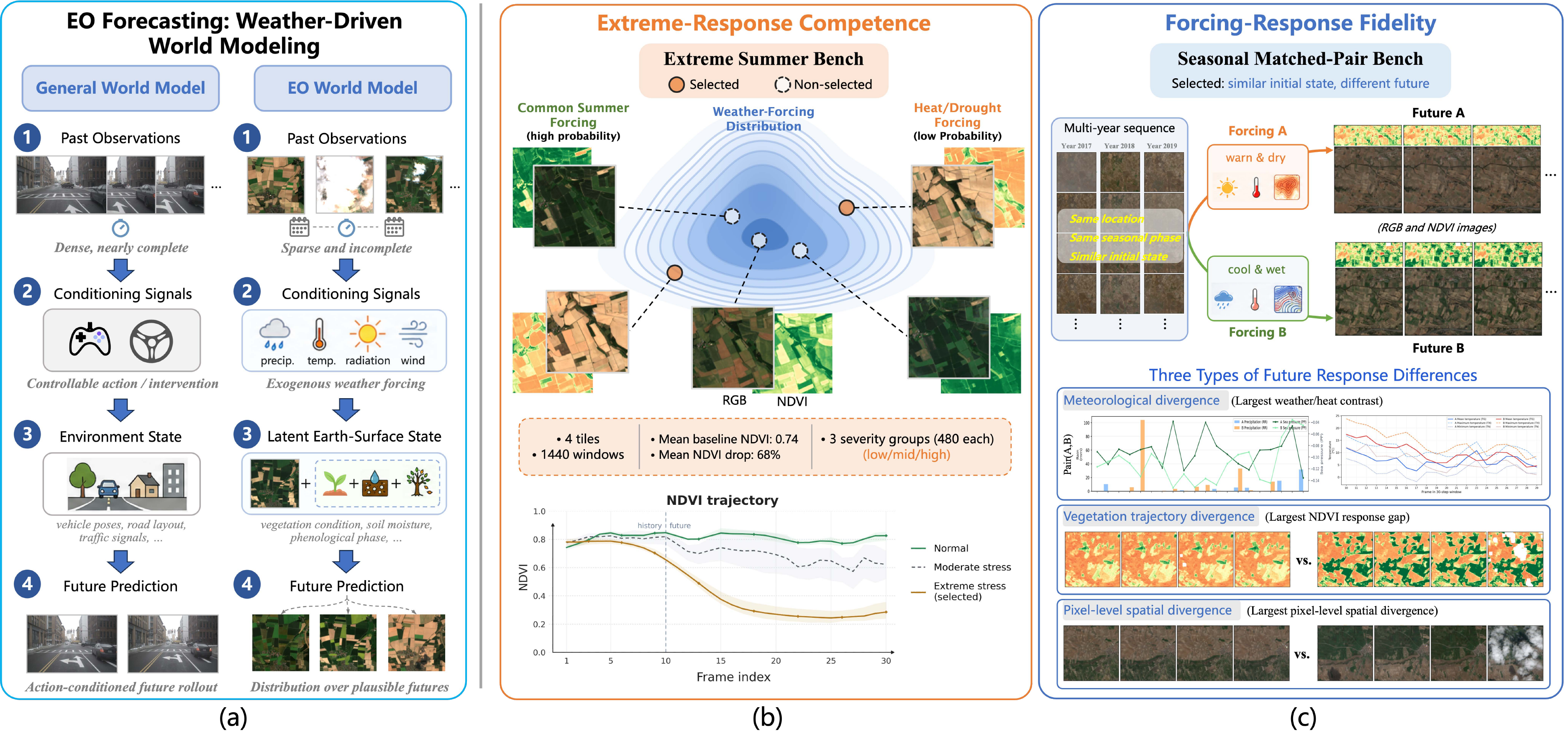}
    \caption{\textbf{Overview of EO world model and the proposed evaluation benchmarks.} (a): EO forecasting differs from standard action-conditioned world modeling: satellite observations are sparse and incomplete, and exogenous weather forcing drives future surface change in ways that depend on unobserved latent Earth-surface states.
    (b) and (c): We define two EO-specific evaluation dimensions beyond standard reconstruction: predicting vegetation degradation under drought and heat stress, and preserving the correct surface response when meteorological forcing changes.
    }
    \label{fig:intro}
    \vspace{-15pt}
\end{figure}

However, as shown in Fig.~\ref{fig:intro} (a), EO forecasting differs in two fundamental ways from the standard world-modeling settings: (1) EO observations are sparse and incomplete. Satellite revisit intervals are on the order of days, and cloud contamination further reduces the frequency of valid observations, leaving the dynamics between observations unobserved. (2) Although meteorology acts as an exogenous condition of surface state transitions, the forcing-response mapping remains stochastic under partial observability, as similar weather forcings can still lead to different surface outcomes because of unobserved internal variability and latent land-surface states (e.g., soil moisture)~\cite{deser2020insights}. Together, these characteristics call for probabilistic prediction and for evaluation that goes beyond visual fidelity to test whether a model responds faithfully to changes in exogenous forcing.

Existing EO forecasting methods address only parts of this picture. Deterministic predictors~\cite{gao2022earthformer,benson2024multi,Shinohara_2025_vitkoop} provide strong point predictions, but cannot explicitly represent predictive uncertainty. More recent diffusion-based methods~\cite{zhao2025vegediff} move toward probabilistic forecasting. However, meteorological variables are still largely used as generic conditioning signals, without distinguishing among climatological background, anomalous weather events, and accumulated environmental stress. Meanwhile, existing benchmarks~\cite{requena2021earthnet2021,benson2024multi} evaluate agreement with the realized future using pixel-level reconstruction and Normalized Difference Vegetation Index (NDVI) temporal consistency. However, they do not explicitly test whether a model produces physically consistent surface responses when meteorological forcing changes. Such forcing-response capability is central to world modeling, where a model is expected to simulate how the world evolves under different actions or external conditions~\cite{tian2023vp2,shang2026worldarena,xu2026worldmark}.

To address these limitations, we present EO-WM, a video diffusion transformer for EO world modeling. EO-WM predicts future multispectral satellite imagery from past sparse observations and heterogeneous EO conditions. This diffusion-based formulation allows the model to represent unobserved intermediate dynamics and multiple plausible futures under sparse, partial observability. Meanwhile, weather acts as the key observable condition signal of Earth surface dynamics under partial observability, but its effect is not well captured when weather variables are treated as undifferentiated conditioning channels. Therefore, we introduce a physically informed conditioning framework based on the physical structure of meteorological forcing. Specifically, we decompose weather forcing into a climatological baseline and an anomalous component, and inject them through separate conditioning pathways according to their distinct physical roles.
We further accumulate anomalous forcing over time into cumulative stress indices, which capture how long abnormal weather persists and help the model distinguish short-lived fluctuations from sustained heat and drought.

Furthermore, we introduce two benchmarks built on EarthNet2021~\cite{requena2021earthnet2021} to evaluate capabilities that are not captured by standard pixel-level metrics, as in Fig.~\ref{fig:intro} (b) and (c). The \textit{Extreme Summer Benchmark} measures whether a model can predict vegetation degradation under drought and heat stress. The \textit{Seasonal Matched-Pair Benchmark} pairs the same geographic location under different weather conditions across years, testing whether predictions change in the correct direction and with a proportionate magnitude when the forcing changes. Our main contributions are as follows:
{
\begin{itemize}[leftmargin=18pt,itemsep=1pt,topsep=2pt,parsep=0pt,partopsep=0pt]
    \item We frame Earth Observation forecasting as a partially observed, weather-driven world modeling problem
    and propose EO-WM, a video diffusion transformer for probabilistic EO forecasting. EO-WM introduces a physically informed conditioning framework that decomposes meteorological forcing into climatological baseline, weather anomaly, and cumulative stress signals based on their physical roles.

    \item We introduce two benchmarks: Extreme Summer and Seasonal Matched-Pair. The former evaluates vegetation degradation onset and severity under rare heat and drought forcing, and the latter tests whether predictions respond in direction and magnitude under changed forcing.
    
    \item Experiments show that EO-WM achieves stronger weather-response fidelity than deterministic models and diffusion-based models, with higher extreme-event detection and forcing-response fidelity while remaining competitive on standard reconstruction metrics.
\end{itemize}
}

\section{Related Work}
\subsection{Video Prediction, Controllable Generation, and World Models}

Diffusion-based video prediction provides a natural way to model uncertain future dynamics~\cite{hoppe2022diffusion,luvdt,zhang2024extdm,voleti2022mcvd,ye2024stdiff,pallotta2025syncvp}. Recent controllable video generation models have further improved how external conditions are injected into large video backbones~\cite{ma2025controllable}. Methods such as STIV~\cite{lin2025stiv} and ATI~\cite{wang2025ati} incorporate text, image, trajectory, or motion controls into diffusion transformers, while large-scale systems such as Wan~\cite{wan2025wan} and Open-Sora~\cite{zheng2025open} demonstrate strong controllable generation quality with flexible conditioning interfaces. These conditions are mainly used to specify semantic content, motion patterns, or camera behavior, and are typically evaluated by alignment with the given control. In EO forecasting, meteorological inputs are observed exogenous forcings, and the key question is whether the predicted surface change responds to these forcings in a physically consistent way.

A recent line of video-based world modeling work has moved beyond passive generation toward interactive simulation. One direction develops large world foundation models or open interactive simulators~\cite{bruce2024genie,agarwal2025cosmos,team2026advancing}. Another direction adapts pretrained video diffusion models into action-conditioned or interactive world models~\cite{rigter2024avid,huang2025vid2world}. These works show that video generation models can be extended toward world simulation when paired with actions or controls. However, most of their conditioning signals are typically an agent action, camera motion, or user control. EO forecasting differs from this setting: the signal is observed meteorological forcing rather than a controllable action. This motivates EO-specific probabilistic modeling and evaluation protocols that test forcing-response fidelity beyond visual reconstruction.

\subsection{Earth Observation Forecasting and Generative Models}

EO forecasting has been formalized by EarthNet2021~\cite{requena2021earthnet2021} as predicting future satellite observations conditioned on the weather. Early methods mainly adopt deterministic prediction. ConvLSTM-based models~\cite{diaconu2022understanding} show that explicit weather conditioning improves forecasting performance, Earthformer~\cite{gao2022earthformer} provides a strong spatiotemporal transformer backbone, and vegetation forecasting studies highlight the importance of multi-modal context and weather signals~\cite{benson2024multi,kladny2024enhanced,janetzky2024global}. These methods establish EO forecasting as a weather-guided prediction task, but they typically produce a single forecast and therefore cannot explicitly represent predictive uncertainty under partial observability.

Generative EO models have recently started to address this limitation. Some methods~\cite{smith2024earthpt,stock2024diffobs,zhao2025vegediff,shu2025restore} study diffusion-based satellite forecasting and reconstruction, while UniTS~\cite{zhang2025units} explores a unified framework for remote-sensing time-series tasks. Meanwhile, related generative models~\cite{khanna2023diffusionsat,zheng2024changen2,goktepe2025ecomapper} show the potential of metadata-conditioned and climate-aware generation. Recently, RemoteBAGEL~\cite{lu2025remote} and RS-WorldModel~\cite{xu2026rs} extend foundation-model and world-modeling concepts to remote sensing. However, these methods mainly focus on generic generative quality. They do not explicitly structure weather into climatological baseline, anomaly, and accumulated stress, nor do they evaluate whether generated futures respond faithfully to changes in meteorological forcing.

\section{Method}

\vspace{-5pt}

We formalize the EO forecasting task as weather-driven world modeling under partial observability~(Sec.\ref{sec:prelim}), then describe EO-WM's architecture and generative formulation~(Sec.\ref{sec:arch}). We then present our two physically informed conditioning designs: climatology-anomaly decomposition of weather forcing~(Sec.\ref{sec:clim_anom}) and cumulative physical stress conditioning~(Sec.\ref{sec:cumstress}).

\subsection{Problem Formulation}
\label{sec:prelim}
We consider an EO forecasting task defined over multispectral satellite image sequences. Let $\mathbf{o}_i \in \mathbb{R}^{C \times H \times W}$ denote the satellite observation at sparse timestamp $u_i$, where $i=1,\ldots,T$ indexes satellite frames and $C$ is the number of spectral bands. Let $\mathbf{a}_\ell \in \mathbb{R}^{C_a \times H_a \times W_a}$ denote dense meteorological forcing at time step $\ell=1,\ldots,L$ on the weather grid, where $C_a$ is the number of meteorological channels (five in EarthNet2021). We use $\pi(i)\in\{1,\ldots,L\}$ to denote the dense-weather time index aligned with satellite frame $i$. Given past sparse observations $\mathbf{o}_{1:T_\text{in}}$, the goal is to predict future observations $\mathbf{o}_{T_\text{in}+1:T}$, conditioned on dense meteorological forcing $\mathbf{a}_{1:L}$, static geographic context $\mathbf{s}$ (e.g., Digital Elevation Model, DEM), and spatiotemporal metadata $\mathbf{m}$ (e.g., location and calendar time). The generative objective is:
\begin{equation}
    p_\theta\!\left(\mathbf{o}_{T_\text{in}+1:T} \mid \mathbf{o}_{1:T_\text{in}},\; \mathbf{a}_{1:L},\; \mathbf{s},\; \mathbf{m}\right).
    \label{eq:objective}
\end{equation}
Text captions are used only as auxiliary backbone context and are not part of the core task definition. This formulation treats weather as an exogenous driver of surface-state transitions, analogous to the conditioning signal in action-conditioned world models~\cite{ha2018world,hafner2019learning}. At the same time, the mapping from $(\mathbf{o}_{1:T_\text{in}}, \mathbf{a}_{1:L})$ to $\mathbf{o}_{T_\text{in}+1:T}$ is not deterministic: the same weather forcing can produce different surface outcomes depending on unobserved land-surface conditions. This motivates a probabilistic generative model that can preserve observed context while representing multiple plausible futures.

\subsection{Model Architecture}
\label{sec:arch}

\begin{figure}[!t]
    \centering
    \includegraphics[width=0.99\textwidth]{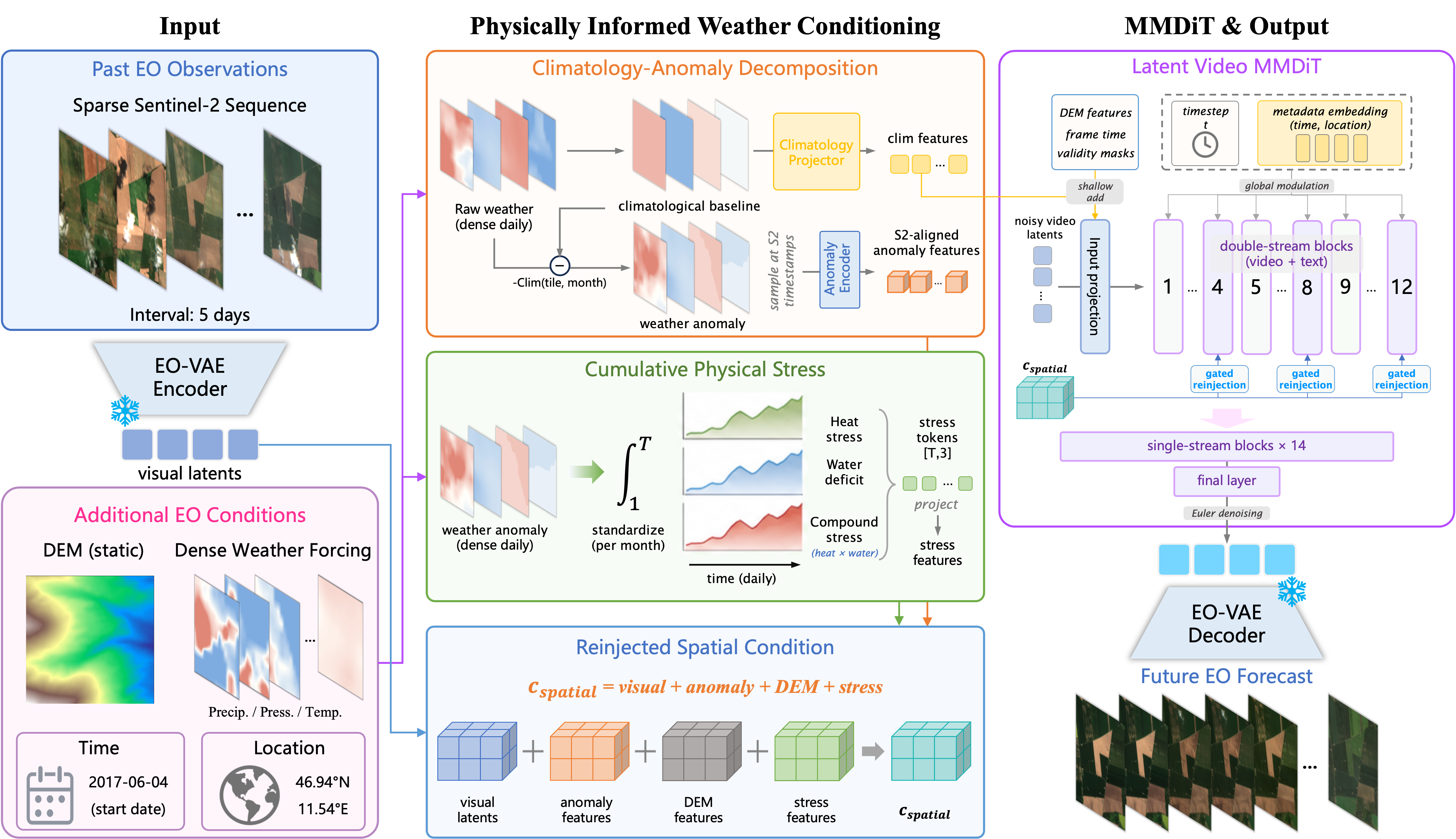}
    \caption{\textbf{Overview of EO-WM.} Sparse EO observations are encoded into visual latents, while dense daily weather forcing is decomposed using $\mathrm{Clim}(\text{tile}, \text{month})$, a precomputed monthly climatology for each geographic tile. The climatological features are injected through a shallow conditioning path, whereas anomaly, DEM, visual, and cumulative-stress features are combined into a spatial condition.
}
    \label{fig:pipeline}
    \vspace{-8pt}
\end{figure}

To model stochastic futures under sparse observations and exogenous conditions, EO-WM is built on a latent diffusion architecture~\cite{zheng2025open}. As shown in Fig.~\ref{fig:pipeline}, an EO-specific variational autoencoder (VAE)~\cite{lehmann2026eo} encodes the multispectral input $\mathbf{o}_{1:T} \in \mathbb{R}^{C \times T \times H \times W}$ into a clean latent representation $\mathbf{z}_0 \in \mathbb{R}^{D \times T \times H' \times W'}$, where $D$ is the latent dimension and $H', W'$ are the downsampled spatial sizes.
The core generative model is a Multimodal Diffusion Transformer (MMDiT) trained with flow matching~\cite{lipman2022flow}. Let $\mathbf{c}$ collect all
conditioning inputs, including mask-aware observations, meteorological
forcing, static geographic context, metadata, and auxiliary text embeddings. Given $\mathbf{z}_0$ and Gaussian noise
$\boldsymbol{\epsilon}\sim\mathcal{N}(\mathbf{0},\mathbf{I})$, we sample a shifted
flow time $r\in(0,1)$ and form the noisy latent $\mathbf{z}_r$ with target velocity $\mathbf{v}^{\star}$:
\begin{align}
    \mathbf{z}_r
    &= (1-r)\mathbf{z}_0 +
    \left[\sigma_{\min} + (1-\sigma_{\min})r\right]\boldsymbol{\epsilon},
    \label{eq:fm_path} \\
    \mathbf{v}^{\star}
    &= \frac{d\mathbf{z}_r}{dr}
    = (1-\sigma_{\min})\boldsymbol{\epsilon}-\mathbf{z}_0,
    \label{eq:fm_target}
\end{align}
where $\sigma_{\min}$ is a small minimum-noise constant. The MMDiT predicts $\mathbf{v}_\theta(\mathbf{z}_r,r;\mathbf{c})$ and is trained to match $\mathbf{v}^{\star}$ with mean-squared error, excluding conditioned context frames and invalid pixels.

\paragraph{Multi-source condition routing.} As summarized in Fig.~\ref{fig:pipeline}, EO-WM routes heterogeneous conditions by their roles. The noisy latent video $\mathbf{z}_r$ forms the main video-token stream. Timestep, geospatial metadata, and auxiliary captions use the inherited MMDiT conditioning interfaces. Climatology features are added once at the input token layer as a seasonal reference, while spatially aligned conditions are collected into a reinjected condition $\mathbf{c}_\text{spatial}$:
\begin{equation}
    \mathbf{c}_\text{spatial}
    =
    \mathbf{c}_\text{vis}
    + \mathbf{c}_\text{dem}
    + \mathbf{c}_\text{time}
    + \mathbf{c}_\text{anom}
    + \mathbf{c}_\text{stress},
    \label{eq:cspatial}
\end{equation}
where $\mathbf{c}_\text{vis}$ is the projected mask-aware visual context, $\mathbf{c}_\text{dem}$ is the DEM features, $\mathbf{c}_\text{time}$ is the frame-time embedding, and $\mathbf{c}_\text{anom}$ and $\mathbf{c}_\text{stress}$ are produced by the weather modules below. All terms share the packed video-token layout, so $\mathbf{c}_\text{spatial}$ can be added directly to video-token hidden states during reinjection.
Spatial conditions injected only at the input may weaken as features propagate through the transformer. Therefore, EO-WM periodically reinjects $\mathbf{c}_\text{spatial}$ into the video-token stream after selected double-stream blocks using zero-initialized learned gates. In the final model, reinjection occurs every four double-stream blocks. This lightweight mechanism keeps observation-aware and forcing-aware spatial signals available at depth, while the main physical innovations remain the anomaly and cumulative-stress condition designs introduced next.

\subsection{Climatology--Anomaly Decomposition of Weather Forcing}
\label{sec:clim_anom}

Absolute meteorological values mix two different sources of physical information: a slowly varying seasonal background and a departure from that background. Thus, we decompose weather forcing into a climatological baseline and a residual anomaly, then route the two components through pathways that match their physical roles. For each geographic tile $q$ and calendar month $m$, we precompute a monthly climatological mean $\bar{\mathbf{a}}_{q,m}\in\mathbb{R}^{C_a}$. For satellite frame $i$, with aligned dense-weather index $\ell_i=\pi(i)$ and calendar month $m_i$, we define:
\begin{equation}
    \mathbf{a}_i^{\text{clim}} = \bar{\mathbf{a}}_{q,m_i}, \qquad
    \mathbf{a}_i^{\text{anom}} = \mathbf{a}_{\ell_i} - \bar{\mathbf{a}}_{q,m_i},
    \label{eq:clim_anom}
\end{equation}
where $\mathbf{a}_i^{\text{clim}}$ is the satellite-aligned climatological baseline and $\mathbf{a}_i^{\text{anom}}$ is the residual anomaly. The baseline is a compact seasonal reference and is added once to the input token stream, while the anomaly retains its spatial field structure, is sampled to align Sentinel-2 timestamps, and enters the reinjected spatial pathway as $\mathbf{c}_\text{anom}$. This separation lets the model condition on the expected seasonal regime without treating normal seasonal variation as active forcing.

This design also supports anomaly-targeted classifier-free guidance (CFG)~\cite{ho2021classifier}: during training, we randomly drop the anomaly tensor while retaining climatology, and at inference, we can compare full and no-anomaly predictions to amplify sensitivity to unusual forcing, as shown in Tab.~\ref{tab:inference_strategy_ablation}.

\subsection{Cumulative Physical Stress Conditioning}
\label{sec:cumstress}

The anomaly decomposition above captures instantaneous departures from climatology, but land-surface response often depends on the integrated history of forcing. For example, the vegetation degradation under heat or drought typically emerges after sustained exposure rather than isolated spikes. Therefore, EO-WM derives cumulative stress features on the dense meteorological timeline and samples them at the sparse satellite observation dates.

We first standardize anomaly fields by the per-tile monthly climatological standard deviation, using $\tilde{a}$ to denote the resulting standardized anomaly. Let $\tilde{a}^{\text{temp}}_\ell(\mathbf{x})$ and $\tilde{a}^{\text{precip}}_\ell(\mathbf{x})$ denote the temperature and precipitation components at dense weather time $\ell$ and weather-grid location $\mathbf{x}$. Specifically, we accumulate three stress fields:
\begin{align}
    \text{Heat stress:} \quad
    & S_\ell^{\text{heat}}(\mathbf{x})
    = \textstyle\sum_{\tau=1}^{\ell}
    \mathrm{ReLU}\!\left(\tilde{a}_\tau^{\text{temp}}(\mathbf{x})\right),
    \label{eq:heat} \\
    \text{Water deficit:} \quad
    & S_\ell^{\text{water}}(\mathbf{x})
    = \textstyle\sum_{\tau=1}^{\ell}
    \mathrm{ReLU}\!\left(-\tilde{a}_\tau^{\text{precip}}(\mathbf{x})\right),
    \label{eq:water} \\
    \text{Compound stress:} \quad
    & S_\ell^{\text{comp}}(\mathbf{x})
    = S_\ell^{\text{heat}}(\mathbf{x}) \cdot S_\ell^{\text{water}}(\mathbf{x}).
    \label{eq:compound}
\end{align}
The ReLU gates keep only the harmful direction of each anomaly: positive temperature anomalies for heat stress and negative precipitation anomalies for water deficit. At each satellite-aligned time, we spatially average and log-compress the three stress values, project the resulting stress tokens, and add them to the reinjected spatial condition as $\mathbf{c}_\text{stress}$. Thus, EO-WM receives both instantaneous forcing departures and their accumulated burden without changing the diffusion objective.

\vspace{-5pt}

\section{Benchmarks and Metrics}
\label{sec:benchmarks}

Standard EarthNet2021 metrics~\cite{requena2021earthnet2021} ask whether a prediction matches one realized future, but they do not isolate whether a model has learned the weather-driven transition behavior expected from an EO world model. We therefore build two diagnostic benchmarks from the EarthNet2021 test splits. Tab.~\ref{tab:metrics} summarizes how each benchmark is constructed, what forecasting behavior it probes, and which metrics we report. Detailed pipelines and metrics are given in the appendix.

\begin{table}[!h]
\centering
\caption{\textbf{Benchmark construction pipeline and metric design.} Both benchmarks retain the EarthNet2021 10-context/20-target forecasting protocol, but focus on model behaviors that standard reconstruction metrics do not capture.}
\label{tab:metrics}
\small
\setlength{\tabcolsep}{3pt}
\begin{tabular}{p{0.12\linewidth}p{0.38\linewidth}p{0.25\linewidth}p{0.17\linewidth}}
\toprule
\textbf{Benchmark} & \textbf{Construction pipeline} & \textbf{Diagnostic question} & \textbf{Primary metrics} \\
\midrule
\textbf{Extreme Summer} &
\texttt{extreme\_test} $\rightarrow$ NDVI curve smoothing $\rightarrow$ stable-to-decline anchor at the context/target boundary $\rightarrow$ cloud-quality and baseline-relative event verification $\rightarrow$ low/mid/high severity bins. &
Tests whether the model forecasts the \emph{onset and severity} of vegetation degradation from a healthy context under heat/drought forcing. &
\textbf{Standard:} ENS/P-MAE/N-MAE \newline
\textbf{Diagnostic:} TN-MAE and DAE by severity bin \\
\midrule
\textbf{Seasonal Matched-Pair} &
\texttt{seasonal\_test} $\rightarrow$ same-location/same-season cross-year pairs $\rightarrow$ quality filtering and initial-state matching $\rightarrow$ stratified selection over meteorological, vegetation-trajectory, and pixel-spatial divergence. &
Tests whether changing weather forcing produces the correct \emph{direction, magnitude, and ranking} of future vegetation divergence. &
\textbf{Standard:} ENS/P-MAE/N-MAE \newline
\textbf{Diagnostic:} DRR, DHR, and PDC \\
\bottomrule
\end{tabular}
\end{table}

\paragraph{Extreme Summer Benchmark.}
The Extreme Summer Benchmark contains 1{,}440 verified windows from the 2018 European summer heat event. We use NDVI trajectory analysis to place each 30-frame window so that the 10-frame context ends immediately before a vegetation decline, then require valid cloud masks and a baseline-relative NDVI drop in the 20-frame target period. We split the verified windows into low-, mid-, and high-severity bins according to the NDVI decline amplitude, so the evaluation can expose failures that are concentrated on stronger events. Therefore, this benchmark tests a precise function: given a healthy observed context and future heat/drought forcing, can the model predict when vegetation degrades and how severe the decline is?

\vspace{-3pt}

\paragraph{Seasonal Matched-Pair Benchmark.}
The Seasonal Matched-Pair Benchmark contains 422 pairs from 380 locations in the full seasonal-cycle test set. Each pair comes from the same geographic cube and seasonal timing, but from different years. We further apply quality filtering and initial-state matching to reduce cloud, phenology, and observed-context confounds, then select pairs through three complementary tracks: meteorological divergence, vegetation-trajectory divergence, and pixel-level spatial divergence. This benchmark tests a different function: under matched location and initial state, does changing the weather forcing change the predicted vegetation future in the same direction and with comparable strength as the real world?

\vspace{-3pt}

\paragraph{Metrics.}
We report standard reconstruction metrics for context: EarthNetScore (ENS), the official EarthNet2021 aggregate score combining MAD, OLS, EMD, and SSIM sub-scores; Pixel-MAE (P-MAE); and NDVI-MAE (N-MAE). For Extreme Summer, Trough NDVI-MAE (TN-MAE) measures NDVI error at the ground-truth trough, and Drop Amplitude Error (DAE) measures error in baseline-to-trough NDVI decline amplitude.
For Seasonal Matched-Pair, Divergence Reproduction Ratio (DRR) compares predicted and ground-truth absolute divergence magnitudes and is best when close to 1; Directional Hit Rate (DHR) measures the sign accuracy of pairwise NDVI differences on sufficiently divergent target frames; and Paired Divergence Correlation (PDC) is the Spearman correlation between predicted and ground-truth per-pair total absolute divergence. These metrics separately measure response magnitude, direction, and ranking.

\section{Experiments}

\begin{table*}[!t]
\centering
\caption{\textbf{Comparison of deterministic and generative models on the Extreme Summer Benchmark.} We report TN-MAE and DAE under low-, mid-, and high-intensity extreme event bins. ``*'' indicates that the model is finetuned from its pretrained weight.}
\label{tab:model_comparison}
\footnotesize
\setlength{\tabcolsep}{3pt}
\begin{tabular}{l|ccc|cc|cc|cc}
\toprule
\multirow{2}{*}{\textbf{Model}} 
& \multirow{2}{*}{\textbf{ENS}$\uparrow$}
& \multirow{2}{*}{\textbf{P-MAE}$\downarrow$}
& \multirow{2}{*}{\textbf{N-MAE}$\downarrow$}
& \multicolumn{2}{c|}{\textbf{Low bin}}
& \multicolumn{2}{c|}{\textbf{Mid bin}}
& \multicolumn{2}{c}{\textbf{High bin}} \\
\cmidrule(lr){5-6} \cmidrule(lr){7-8} \cmidrule(lr){9-10}
&  &  &  
& \textbf{TN-MAE}$\downarrow$ & \textbf{DAE}$\downarrow$
& \textbf{TN-MAE}$\downarrow$ & \textbf{DAE}$\downarrow$
& \textbf{TN-MAE}$\downarrow$ & \textbf{DAE}$\downarrow$ \\
\midrule
\multicolumn{10}{l}{\textbf{Deterministic Models}} \\
\midrule
SimVP~\cite{gao2022simvp}                 & 0.1177 & 0.0686 & 0.1846 &      0.2142 &0.2857 & 0.2275 & 0.2844 & 0.2247 & 0.2896  \\
PhyDNet~\cite{guen2020disentangling}      & 0.1157 & 0.0918 & 0.2639 &   0.2733 & 0.3317 & 0.2761 & 0.3464 & 0.2458 & 0.3458  \\
PredRNN~\cite{wang2017predrnn}            & 0.0589 & 0.1357   & 0.3757  & 0.3782 & 0.3169 & 0.3812 & 0.3226 & 0.3721 & 0.3154  \\
PredRNN\,v2~\cite{wang2022predrnn}         & 0.1435  & 0.0655  & 0.2358  & 0.2643 & 0.2961 & 0.2874 & 0.3098 & 0.2749 & 0.3073  \\
TAU~\cite{tan2023temporal}           & 0.1348 & 0.0781 & 0.2293  & 0.2456 &	0.3229 & 0.2553 & 0.3525 & 0.2441 & 0.3488   \\
Earthformer~\cite{gao2022earthformer}     & 0.2539 & 0.0355 & \textbf{0.1099} 
& 0.1271 & 0.2227 
& 0.1348 & 0.2801 
& 0.1338 & 0.3084 \\
\midrule
\multicolumn{10}{l}{\textbf{Generative Models}} \\
\midrule
Wan2.1-Fun-InP*~\cite{wan2025wan}         & 0.2028 & 0.0434 & 0.1345 
& 0.1371 & \textbf{0.2176} 
& 0.1447 & 0.2538 
& 0.1426 & 0.2694 \\
Latte~\cite{ma2024latte}                  & 0.1480 & 0.0581  & 0.1983  & 0.1934  & 0.2342 & 0.1972 & 0.2868 & 0.2020 & 0.3017\\
EO-WM (Ours)                              & \textbf{0.2543} & \textbf{0.0332} & \underline{0.1106} 
& \textbf{0.1266} & \underline{0.2192}
& \textbf{0.1296} & \textbf{0.2427} 
& \textbf{0.1281} & \textbf{0.2372} \\
\bottomrule
\end{tabular}
\vspace{-5pt}
\end{table*}

\begin{table}[!t]
\centering
\caption{\textbf{Comparison of deterministic and generative models on the Seasonal Matched-Pair Benchmark.} ``*'' indicates that the model is finetuned from its pretrained weight. DRR$_\mathrm{mean}$: mean value of DRR. PDC$_\mathrm{sp}$: Spearman paired divergence correlation.}
\label{tab:model_comparison_seasonal}
\small
\setlength{\tabcolsep}{3pt}
\begin{tabular}{l|cccccc}
\toprule
\textbf{Model} 
& \textbf{ENS}$\uparrow$ 
& \textbf{Pixel-MAE}$\downarrow$ 
& \textbf{NDVI-MAE}$\downarrow$ 
& \textbf{DRR$_\mathrm{mean}$}$\rightarrow 1$ 
& \textbf{DHR}$\uparrow$ 
& \textbf{PDC$_\mathrm{sp}$}$\uparrow$ \\
\midrule
\multicolumn{7}{l}{\textbf{Deterministic Models}} \\
\midrule
SimVP~\cite{gao2022simvp}        & 0.2098 & 0.0475 & 0.1586 & 1.4467 & 0.4938 & 0.1606 \\
PhyDNet~\cite{guen2020disentangling}      & 0.0626 & 0.1124 & 0.3233 & 1.5278 & 0.5049 & 0.0362 \\
PredRNN~\cite{wang2017predrnn}       & 0.0312  & 0.1325  & 0.3936  & 0.1619  & 0.4436  & -0.1503  \\
PredRNN\,v2~\cite{wang2022predrnn}  &  0.1868  & 0.1070  & 0.3614  & 0.3961  & 0.5049  & 0.1227 \\
TAU~\cite{tan2023temporal}          & 0.1735 & 0.0482 & 0.1512 & 0.5183 & 0.4948 & 0.1386 \\
Earthformer~\cite{gao2022earthformer}  & 0.2879 & 0.0280 & 0.1109 & 0.6302 & 0.5551 & 0.1814 \\
\midrule
\multicolumn{7}{l}{\textbf{Generative Models}} \\
\midrule
Wan2.1-Fun-Inp*~\cite{wan2025wan} & 0.2537 & 0.0325 & 0.1208 & \textbf{0.7066} & 0.6050 & 0.2809 \\
Latte~\cite{ma2024latte}         &  0.2050 & 0.0455 & 0.1680 & 1.4021 & 0.5613 & 0.1882  \\
EO-WM (Ours)   & \textbf{0.3022} & \textbf{0.0222} & \textbf{0.1012} & 
\underline{0.6927} & \textbf{0.6522} & \textbf{0.2942} \\
\bottomrule
\end{tabular}
\vspace{-10pt}
\end{table}

\subsection{Implementation Details}

We follow the standard EarthNet2021 forecasting setting: all methods receive 10 context frames and the same weather conditions, and predict 20 future 4-channel Sentinel-2 frames at $128\times128$ resolution. For EO-WM, the EO-VAE tokenizer~\cite{lehmann2026eo} is finetuned on the EarthNet2021 training split. The diffusion backbone is trained from scratch with 387M parameters. Unless otherwise specified, EO-WM does not use CFG during inference. More details are in the appendix.

For comparison methods, all methods except Wan2.1-Fun-V1.1-1.3B-InP (abbreviated as Wan2.1) are trained on the same EarthNet2021 training split. Earthformer follows the official implementation. Latte and the OpenSTL~\cite{tan2023openstl} baselines (SimVP, TAU, PredRNN, PredRNNv2, and PhyDNet) are adapted to 4-channel EO input/output protocol and receive the same weather variables through cross-attention and FiLM-based~\cite{perez2018film} conditioning, respectively. Wan2.1 is initialized from the official 1.3B checkpoint and is adapted through a carefully designed four-stage fine-tuning procedure. For stochastic generative models, we draw five predictions and evaluate the ensemble mean unless otherwise specified. More details are provided in the appendix.

\vspace{-5pt}

\subsection{Comparisons}

Tables~\ref{tab:model_comparison} and~\ref{tab:model_comparison_seasonal} show that standard reconstruction quality alone does not fully capture the desired EO world-modeling behavior. Earthformer remains a strong deterministic baseline and gives the lowest overall NDVI-MAE on Extreme Summer, but its drop-amplitude error increases with event severity, indicating conservative forecasts that under-reproduce large vegetation declines. Generative baselines provide stochastic futures. Wan2.1 benefits from powerful pretrained spatiotemporal dynamics, but generic video priors alone do not consistently preserve EO calibration or weather-response direction. EO-WM combines competitive pixel fidelity with the strongest event-severity and paired-response metrics: it achieves the best TN-MAE in all severity bins on Extreme Summer and the best DHR/PDC on Seasonal Matched-Pair, indicating more faithful forcing-conditioned surface dynamics.

\vspace{-5pt}

\subsection{Ablation Studies}

\begin{table*}[!t]
\centering
\caption{\textbf{Ablation of physically informed weather conditions.} ``Weather repr.'' denotes the representation of weather conditions. The first row feeds raw weather conditions through the same learned spatial encoder and reinjection pathway. \emph{Decomp}: climatology--anomaly decomposition; \emph{CumSt}: cumulative stress conditioning. The evaluation uses no inference-time CFG.}
\label{tab:physical_design_ablation}
\small
\setlength{\tabcolsep}{3pt}
\begin{tabular}{l|cccc|cccc}
\toprule
\multirow{2}{*}{Weather repr.} & \multicolumn{4}{c|}{Extreme Summer} & \multicolumn{4}{c}{Seasonal Matched-Pair} \\
& ENS$\uparrow$ & N-MAE$\downarrow$ & TN-MAE$\downarrow$ & DAE$\downarrow$ & ENS$\uparrow$ & DRR$_\mathrm{mean}$ $\rightarrow 1$ & DHR$\uparrow$ & PDC$_\mathrm{sp}$$\uparrow$ \\
\midrule
Raw & 0.2538& 0.1114 &  0.1187 & 0.2459 &  0.3007 &  0.6515 &  0.6127 & 0.2704  \\
\emph{Decomp} &  0.2541 & 0.1118 & \textbf{0.1171} & 0.2367 & 0.3012 & \textbf{0.6939} & 0.6247 & 0.2919 \\
\emph{Decomp} + \emph{CumSt} & \textbf{0.2543} & \textbf{0.1106} & 0.1179 & \textbf{0.2330} & \textbf{0.3022} & 0.6927 & \textbf{0.6522} & \textbf{0.2942}  \\
\bottomrule
\end{tabular}
\vspace{-5pt}
\end{table*}

\begin{table*}[!t]
\centering
\caption{\textbf{Ablation of inference strategy.} The same EO-WM checkpoint is evaluated with different ensemble sizes and different anomaly CFG guidance scales ($\lambda_\mathrm{anom}$).}
\label{tab:inference_strategy_ablation}
\small
\setlength{\tabcolsep}{3pt}
\begin{tabular}{cc|cccc|cccc}
\toprule
\multirow{2}{*}{$N$} & \multirow{2}{*}{$\lambda_\mathrm{anom}$} & \multicolumn{4}{c|}{Extreme Summer} & \multicolumn{4}{c}{Seasonal Matched-Pair} \\
& & ENS$\uparrow$ & N-MAE$\downarrow$ & TN-MAE$\downarrow$ & DAE$\downarrow$ & ENS$\uparrow$ & DRR$_\mathrm{mean}$ $\rightarrow 1$ & DHR$\uparrow$ & PDC$_\mathrm{sp}$$\uparrow$ \\
\midrule
1 & 1.0 & 0.2397 & 0.1185 & 0.1264 & 0.2304 & 0.2955 & 0.7085 & 0.6461 & \textbf{0.3059} \\
5 & 1.0 & \textbf{0.2543} & 0.1106 & 0.1179 & 0.2345 & \textbf{0.3022} & 0.6927 & 0.6522 & 0.2942 \\
5 & 2.0 & 0.2539 & \textbf{0.1104} & \textbf{0.1175} & 0.2360 & \textbf{0.3022} & 0.7169 & 0.6538 & 0.2992 \\
5 & 5.0 & 0.2499 & 0.1116 & 0.1198 & 0.2352 & 0.2993 & 0.8173 & \textbf{0.6629} & 0.3035 \\
5 & 10.0 & 0.2401 & 0.1163 & 0.1266 & \textbf{0.2277} & 0.2922 & \textbf{0.9954} & 0.6558 & 0.2972 \\
\bottomrule
\end{tabular}
\vspace{-10pt}
\end{table*}

\textbf{The influence of physically informed forcing representation.}
Tab.~\ref{tab:physical_design_ablation} ablates the two physical conditioning components against the raw-weather control with the same backbone. Climatology--anomaly decomposition mainly improves degradation-amplitude and paired-divergence metrics, consistent with its physical role.
Adding cumulative stress further improves DAE, DHR, and PDC, matching the physical expectation that vegetation response depends not only on instantaneous anomalies but also on sustained heat and water deficit. These gains support the design choice of representing weather by physical role rather than treating it as a single undifferentiated condition.

\textbf{The influence of inference strategy.}
Tab.~\ref{tab:inference_strategy_ablation} isolates the effects of five-sample ensemble averaging and anomaly CFG. Increasing $N$ improves reconstruction-oriented metrics, but slightly reduces PDC, indicating that ensembling stabilizes pixel-level forecasts while damping pair-specific responses. Stronger guidance raises DRR and DHR, but high guidance degrades pixel quality and TN-MAE. $\lambda_\mathrm{anom}=10.0$ shows that DRR$_\mathrm{mean}$ can approach its ideal value through response amplification rather than better ranking or reconstruction. Thus, we use unguided inference for the main architectural comparisons and report anomaly CFG only as an inference-time sensitivity analysis.

\begin{figure}[!t]
    \centering
    \includegraphics[width=0.99\textwidth]{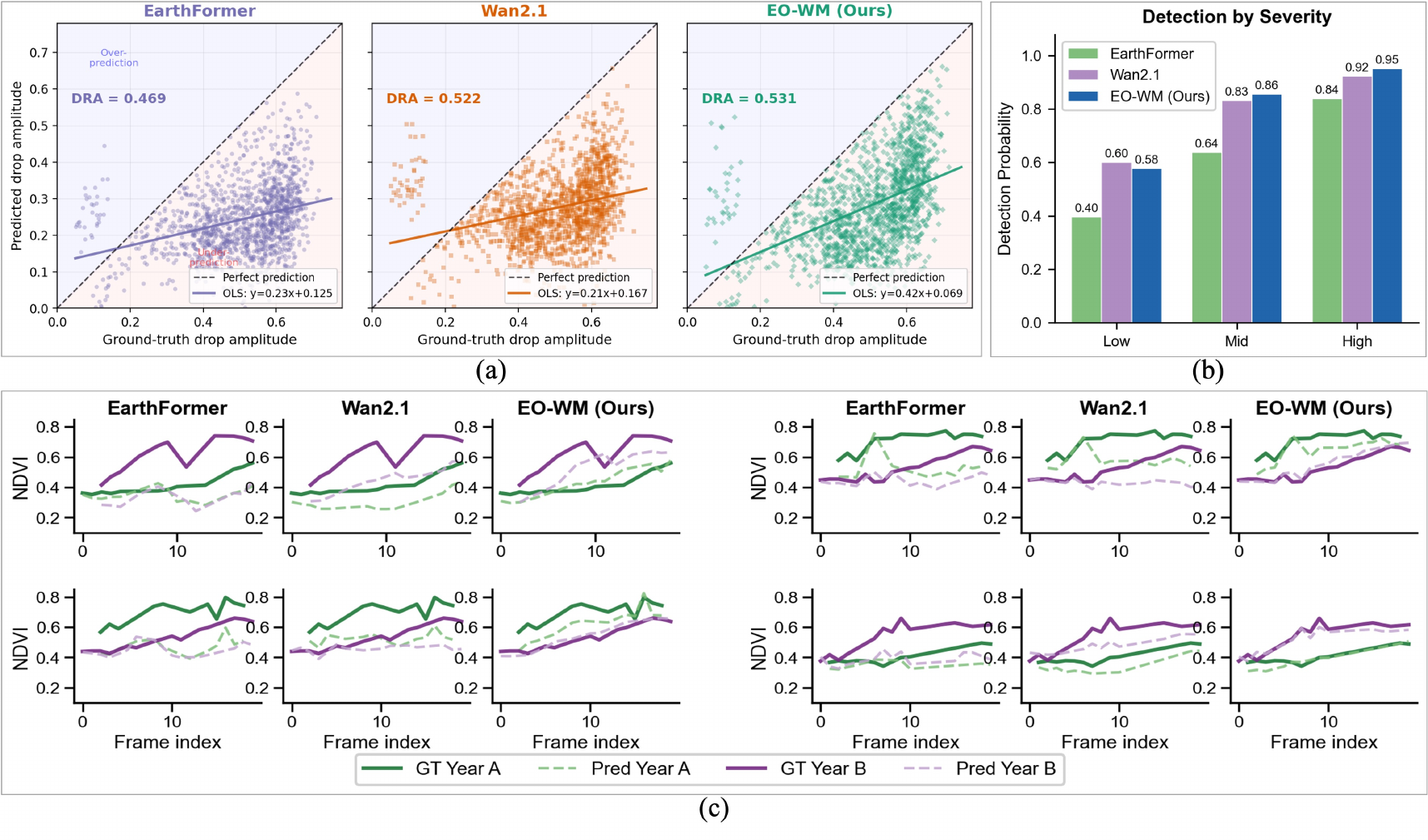}
    \vspace{-3pt}
    \caption{\textbf{Visual diagnostics apart from benchmark metrics.}
    (a)  Predicted versus ground-truth NDVI drop amplitude on the Extreme Summer Benchmark, where the dashed line is perfect severity reproduction. DRA (Drop Reproduction Accuracy) measures relative agreement between predicted and ground-truth drop amplitudes.
    (b) Extreme-event detection rate by severity bin, measured as the fraction of forecasts whose target-period mean NDVI over valid vegetated pixels falls below the benchmark event threshold (0.3).
    (c) Seasonal Matched-Pair NDVI trajectories for same-location cross-year examples. Solid lines show ground truth and dashed lines show predictions.}
    \label{fig:vis_analysis}
    \vspace{-10pt}
\end{figure}

\subsection{Diagnostic Visualization}

Fig.~\ref{fig:vis_analysis} further illustrates the quantitative results by visualizing the behavioral differences behind the benchmark metrics. In Fig.~\ref{fig:vis_analysis} (a), EO-WM has the steepest fitted slope and the highest auxiliary DRA, indicating better severity calibration than others. The detection bars (Fig.~\ref{fig:vis_analysis} (b)) show a complementary advantage of probabilistic forecasting, both generative models detect extreme events much more often than the deterministic EarthFormer, especially in the low- and mid-severity events.
The paired NDVI trajectories show that EO-WM more often preserves the relative ordering and separation between two weather realizations, while the other models less accurately capture the corresponding forcing-response differences.

\vspace{-5pt}

\section{Conclusion}

We presented EO-WM, a physically informed diffusion forecasting model for multispectral Earth Observation forecasting under sparse and partial observations. The model treats meteorology as structured exogenous forcing rather than generic conditioning, separating climatological context, weather anomalies, and cumulative stress. We also introduced two benchmarks that evaluate capabilities beyond pixel reconstruction: predicting degradation under extreme heat and drought, and preserving the correct paired response aligned with observed cross-year weather differences. Across deterministic and generative baselines, EO-WM improves degradation-severity prediction and forcing-response fidelity while remaining competitive on standard pixel-level metrics.

\paragraph{Limitations and broader impact.}
Our current setting can forecast over a seasonal window, but the limited length of paired satellite-observation and weather records makes it difficult to extend directly to multi-year or decadal simulation. Such long-horizon settings would involve hundreds of Sentinel-2 frames, stronger error accumulation, changing seasonal regimes, and slow climate trends. In addition, several land-surface states remain unobserved or only partly observed, including soil moisture, irrigation, and vegetation type.

These limitations also suggest useful directions for future work. For example, combining satellite imagery with ground station measurements collected over the same regions could turn some unobserved hidden states into known conditions. This could improve forecasting accuracy and support ecosystem monitoring, crop-growth prediction, and climate-risk assessment, with positive impacts on society and the environment. Potential negative impacts include overreliance on forecasts for high-stakes agricultural, insurance, or disaster-response decisions.

\bibliographystyle{plainnat}
\bibliography{ref}

\appendix
\numberwithin{equation}{section}
\numberwithin{figure}{section}
\numberwithin{table}{section}


\newpage

\section{Technical appendices and supplementary material}


We organize our supplementary material as follows. Section~\ref{sec:supp_eowm} provides additional EO-WM implementation details, tokenizer reconstruction diagnostics, ablation results, and qualitative examples. Section~\ref{sec:supp_bench} describes the construction of the Extreme Summer and Seasonal Matched-Pair benchmarks, including filtering criteria, sample statistics, and evaluation metrics. Section~\ref{sec:supp_comparison} summarizes the adaptation and training details for the comparison methods, including Wan2.1, Latte and the OpenSTL deterministic baselines. Section~\ref{sec:supp_assets} describes the data and asset availability.

\subsection{EO-WM Training and Ablations}

\label{sec:supp_eowm}


\subsubsection{More Training Details}

EO-WM follows the EarthNet2021 10-to-20 forecasting protocol used in the main paper. Each training example is a 30-frame Sentinel-2 sequence at $128\times128$ resolution with four optical channels (B/G/R/NIR), where the first 10 frames are provided as visual context and the remaining 20 frames are predicted. We first train the EO-VAE on the EarthNet2021 training split, holding out $2\%$ of the training samples for validation. The VAE is trained with base learning rate $2\times10^{-5}$ until convergence. 
We train the MMDiT diffusion backbone from scratch. The backbone uses hidden size 768, 12 attention heads, 12 double-stream blocks, 14 single-stream blocks, patch size 2, and spatial-condition reinjection after every four double-stream blocks.

For data quality control, pixels whose cloud probability is at or above $0.2$ are treated as invalid in the training quality mask. The pixel-space quality mask is downsampled to the latent grid with area averaging: a latent cell is kept in the diffusion loss only when at least $50\%$ of the corresponding pixels are valid. The loss is also masked to exclude the conditioned visual context frames, so optimization is applied to valid target-frame latents rather than to copied reference frames. During training, we randomly drop only the ERA5 anomaly branch with probability $0.15$, while retaining the climatological baseline and other conditioning inputs. At inference time, the no-anomaly prediction used for guidance is obtained by zeroing the anomaly branch.

Auxiliary text captions are generated from a small set of fixed templates using sample metadata and observed sequence descriptors, such as the EarthNet2021 tile, date range, season, frame counts, cloud-condition summary, spectral bands, and the presence of daily meteorological drivers. These captions are used only to populate the inherited text-conditioning interface of the video backbone, they do not include target image content and are not part of the core EO forecasting task definition.

We train on 4 GPUs with bfloat16 mixed precision, ZeRO-2 data-parallel optimization, gradient checkpointing, and no gradient accumulation. For 30-frame clips at $128{\times}128$ resolution, we use batch size 64 per GPU, giving an effective batch size of 256. We optimize with AdamW using learning rate $2\times10^{-4}$, weight decay $0.01$, and cosine learning-rate scheduling with 500 warm-up steps. Flow-matching timestep sampling uses shift parameter $\alpha=2.0$. All EO-WM results reported in this paper are evaluated using the 18{,}000-step checkpoint selected by validation performance.

\subsubsection{Tokenizer Reconstruction Analysis}

We first assess whether the tokenizer reconstruction quality could be a limiting factor for downstream forecasting performance. Table~\ref{tab:vae_recon} compares two fine-tuned tokenizers, EO-VAE and Wan-VAE, against trivial reconstruction references on the two proposed benchmark sets. We report both masked pointwise reconstruction errors (MAE/MSE) and EarthNetScore-style components. This table therefore serves two purposes: it checks whether the tokenizer preserves multispectral observations faithfully, and it illustrates why the main paper reports standard EarthNet-style reconstruction scores together with task-targeted diagnostic metrics.

\paragraph{EarthNetScore components.}
Let $S_{\mathrm{MAD}}$, $S_{\mathrm{OLS}}$, $S_{\mathrm{EMD}}$, and $S_{\mathrm{SSIM}}$ denote the benchmark-level mean subscores after excluding undefined values. Following the EarthNetScore aggregation, the overall score is the harmonic mean of the four components:
\begin{equation}
    \mathrm{ENS}
    =
    H(S_{\mathrm{MAD}}, S_{\mathrm{OLS}}, S_{\mathrm{EMD}}, S_{\mathrm{SSIM}})
    =
    \frac{4}{
    \frac{1}{S_{\mathrm{MAD}}+\epsilon}
    + \frac{1}{S_{\mathrm{OLS}}+\epsilon}
    + \frac{1}{S_{\mathrm{EMD}}+\epsilon}
    + \frac{1}{S_{\mathrm{SSIM}}+\epsilon}},
\end{equation}
where $\epsilon=10^{-8}$. All four components are scaled to $[0,1]$ with higher values indicating better agreement. MAD measures median absolute deviation between predicted and target reflectance values over non-masked pixels and all spectral channels. OLS compares ordinary-least-squares slopes of pixelwise NDVI time series, using non-masked target observations and the corresponding prediction interval. EMD computes a Wasserstein-1 distance between the predicted and observed pixelwise NDVI value distributions, with target distributions formed only from non-masked observations. SSIM computes structural similarity over frames and channels with sufficient valid target pixels, while masked target pixels are filled from the prediction to avoid penalizing unobserved regions. In Table~\ref{tab:vae_recon}, the ENS values in this diagnostic table are the harmonic mean of the EMD, MAD, OLS, and SSIM. We report the unexponentiated SSIM component as Raw-SSIM for interpretability.

\paragraph{Relation to the proposed metrics.}
EarthNetScore is a useful and widely adopted aggregate for EarthNet2021-style forecasting. At the same time, it is not designed to isolate the specific world-modeling behaviors emphasized in the main paper: reproducing the severity of vegetation degradation under heat and drought, and producing appropriately different futures when the meteorological forcing changes. This motivates our evaluation protocol: ENS, P-MAE, and N-MAE are reported as standard reconstruction-oriented context, while TN-MAE and DAE measure event-severity fidelity on Extreme Summer, and DRR, DHR, and PDC measure response magnitude, direction, and ranking on Seasonal Matched-Pair. The reconstruction experiment below provides a controlled example of this complementarity. Even when the target sequence itself is used as a reference, the masked EMD/OLS components need not reach their formal maximum, whereas masked MAE/MSE directly reflect pixelwise reconstruction fidelity. We therefore interpret EarthNet-style scores and the proposed diagnostic metrics as complementary rather than substitutive.

\begin{table*}[!h]
\centering
\caption{\textbf{The performance of tokenizer reconstruction on the two benchmarks.} GT is included as a diagnostic reference, not as a theoretical upper bound for EMD or OLS under masked EarthNetScore-style evaluation. Raw-SSIM denotes the unexponentiated SSIM component used for the harmonic aggregation in this table.}
\label{tab:vae_recon}
\small
\setlength{\tabcolsep}{4pt}
\begin{tabular}{lccccccc}
\toprule
Method & ENS$\uparrow$ & EMD$\uparrow$ & MAD$\uparrow$ & OLS$\uparrow$ & Raw-SSIM$\uparrow$ & MAE$\downarrow$ & MSE$\downarrow$ \\
\midrule
\multicolumn{8}{l}{\textbf{Extreme Summer Benchmark}} \\
\rowcolor{gray!20} Ground Truth & 0.3610 & 0.1661 & 1.0000 & 0.3267 & 1.0000 & 0.0000 & 0.0000 \\
Copy last clear frame & 0.1259 & 0.1361 & 0.1881 & 0.2902 & 0.7660 & 0.0257 & 0.0035 \\
Persistence (cloud-free mean) & 0.1079 & 0.1247 & 0.1849 & 0.2902 & 0.7473 & 0.0263 & 0.0033 \\
EO-VAE (fine-tuned) & 0.2955 & 0.1621 & 0.3127 & 0.3257 & 0.9125 & 0.0022 & 0.0000 \\
Wan-VAE (fine-tuned) & 0.3039 & 0.1999 & 0.2866 & 0.3465 & 0.9454 & 0.0037 & 0.0001 \\
\midrule
\multicolumn{8}{l}{\textbf{Seasonal Matched-Pair Benchmark}} \\
\rowcolor{gray!20} Ground Truth & 0.3809 & 0.1879 & 1.0000 & 0.3332 & 1.0000 & 0.0000 & 0.0000 \\
Copy last clear frame & 0.2790 & 0.2250 & 0.2354 & 0.3333 & 0.9100 & 0.0161 & 0.0013 \\
Persistence (cloud-free mean) & 0.2626 & 0.2137 & 0.2268 & 0.3333 & 0.8950 & 0.0183 & 0.0014 \\
EO-VAE (fine-tuned) & 0.3005 & 0.1735 & 0.3068 & 0.3187 & 0.8694 & 0.0034 & 0.0000 \\
Wan-VAE (fine-tuned) & 0.3048 & 0.2278 & 0.2863 & 0.3407 & 0.9223 & 0.0050 & 0.0001 \\
\bottomrule
\end{tabular}
\end{table*}

\paragraph{Reconstruction fidelity.}
On masked pointwise metrics, EO-VAE reconstruction is near-lossless and clearly stronger than the trivial baselines. On Extreme Summer, EO-VAE reduces MAE from $0.0257$ (copy-last) and $0.0263$ (persistence) to $0.0022$; on Seasonal Matched-Pair, it reduces MAE from $0.0161$ and $0.0183$ to $0.0034$. EO-VAE also achieves lower MAE/MSE than Wan-VAE on both benchmark sets, indicating that the EO-adapted tokenizer preserves EarthNet multispectral observations more faithfully at the pixel level. These results suggest that tokenizer reconstruction is already strong enough that the main bottlenecks in downstream forecasting lie in future-dynamics modeling rather than in the latent autoencoding stage.

\paragraph{Why GT is not the EMD/OLS upper bound.}
A potentially confusing pattern in Table~\ref{tab:vae_recon} is that \textbf{GT does not achieve EMD or OLS equal to $1$}, and some simple or smoothed predictors can even exceed GT on these two submetrics. This is not a computation error, but a consequence of the official EarthNet2021 metric definitions under pixel-level spatiotemporal masking. In the official implementation, EMD is computed for each pixelwise NDVI time series by comparing the prediction distribution over the \emph{full predicted sequence} against the target NDVI distribution formed only from \emph{non-masked target values}. OLS is similarly asymmetric: for each pixelwise NDVI time series, the target slope is fitted only on non-masked target values, whereas the prediction slope is fitted over the contiguous interval between the first and last non-masked target observations. Therefore, exact recovery of the full ground-truth trajectory is not, in general, the optimum for either metric. If the cloud-masked target values differ systematically from the statistics of the non-masked subset, then even the raw GT sequence can score below $1$.

\paragraph{Implication for interpreting the table.}
The table itself shows this asymmetry clearly. On Seasonal Matched-Pair, both Copy last clear frame and Persistence exceed GT in EMD/OLS, even though their MAE/MSE are much worse. Likewise, Wan-VAE achieves higher EMD/OLS than EO-VAE on both benchmark sets despite having consistently worse masked MAE/MSE. These cases indicate that EMD and OLS reward agreement with the \emph{non-masked target-subset statistics of each pixelwise time series} rather than exact pointwise reconstruction of the complete target trajectory. We therefore treat EMD/OLS here as diagnostic references for the behavior of the official EarthNet metrics under sparse, cloud-masked observations, not as tokenizer-fidelity upper bounds. For assessing the reconstruction quality of the tokenizer itself, masked MAE/MSE and Raw-SSIM are the more reliable indicators; under those metrics, EO-VAE reconstruction is substantially stronger than the trivial baselines and slightly better than Wan-VAE.

\subsubsection{Condition-Injection Ablation}

We isolate two design choices: whether all the EO conditions are included in addition to mask-aware visual context, and whether spatial condition features are reinjected inside the MMDiT blocks. The results are shown in Table~\ref{tab:condition_injection_ablation}, the first row uses only visual context and quality masks. The second row adds EO side conditions but injects the resulting features once only at the input. The third row is the default EO-WM setting, with spatial-condition reinjection after every four double-stream blocks. All rows are evaluated without the inference-time CFG.

\begin{table*}[!h]
\centering
\caption{\textbf{Ablation of condition injection.} ``EO cond.'' denotes meteorological, static geographic, and spatiotemporal conditions. ``Deep reinj.'' denotes repeated injection of spatial condition features inside the MMDiT blocks. N-MAE: NDVI-MAE; TN-MAE: trough NDVI-MAE; DAE: drop amplitude error; PDC$_\mathrm{sp}$: Spearman paired divergence correlation.}
\label{tab:condition_injection_ablation}
\small
\setlength{\tabcolsep}{2.2pt}
\begin{tabular}{cc|cccc|cccc}
\toprule
\multirow{2}{*}{EO cond.} & \multirow{2}{*}{Deep reinj.} & \multicolumn{4}{c|}{Extreme Summer} & \multicolumn{4}{c}{Seasonal Matched-Pair} \\
& & ENS$\uparrow$ & N-MAE$\downarrow$ & TN-MAE$\downarrow$ & DAE$\downarrow$ & ENS$\uparrow$ & DRR$_\mathrm{mean}$ $\rightarrow 1$ & DHR$\uparrow$ & PDC$_\mathrm{sp}$$\uparrow$ \\
\midrule
\ding{55} & \ding{55} & 0.1458 & 0.2153 & 0.2140 & 0.2722 & 0.1508 & 0.4524 & 0.4302 & 0.1621 \\
\ding{51} & \ding{55} & 0.2385 & 0.1223 & 0.1286 & 0.2395 & 0.2918 & 0.6612 & 0.6186 & 0.2785 \\
\ding{51} & \ding{51}  & 0.2543 & 0.1106 & 0.1179 & 0.2345 & 0.3022 & 0.6927 & 0.6522 & 0.2942  \\
\bottomrule
\end{tabular}
\end{table*}

As depicted in Table~\ref{tab:condition_injection_ablation}, using only visual context and masks performs poorly on both benchmarks, indicating that visual history alone is insufficient for weather-conditioned EO forecasting. Adding EO side conditions at the input recovers most of the performance, improving Extreme ENS from $0.1458$ to $0.2385$ and Seasonal DHR from $0.4302$ to $0.6186$. Enabling deep spatial-condition reinjection further improves the reconstruction and response-fidelity metrics, suggesting that anomaly and stress signals are more effective when they remain accessible throughout the MMDiT transition layers rather than only at the initial tokenization stage.

\subsubsection{Qualitative Forecasting Results}

\begin{figure*}[!h]
    \centering
    \includegraphics[width=0.99\textwidth]{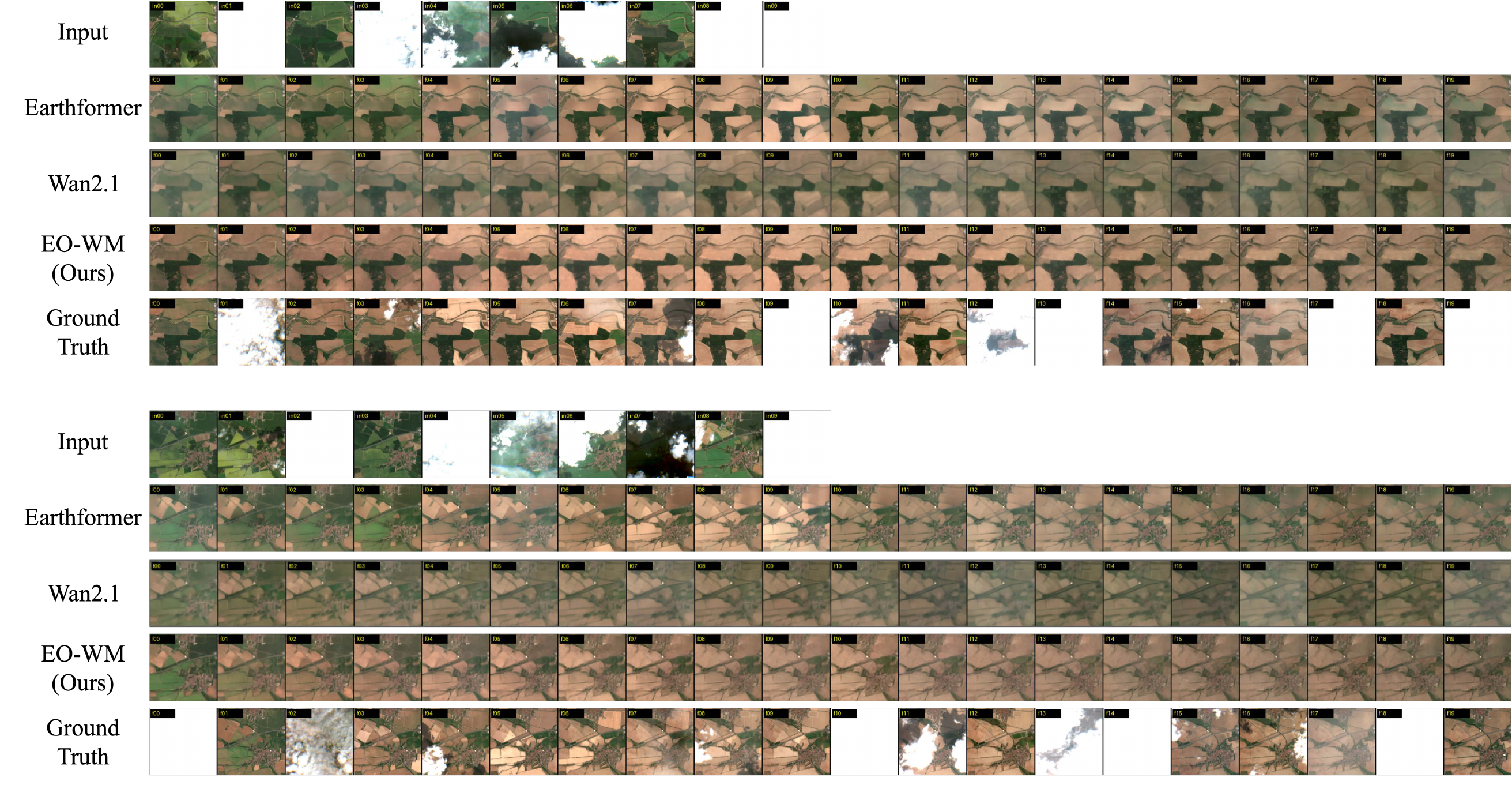}
    \caption{\textbf{Qualitative visualization of 10-to-20 EO forecasting.}
    Each example uses 10 past Sentinel-2 frames as visual context and predicts the next 20 future frames. White regions indicate cloud-contaminated or missing observations in the real Sentinel-2 sequence, illustrating the sparse-observation setting under which the models must infer future surface dynamics.}
    \label{fig:vis_result_comparison}
\end{figure*}

Figure~\ref{fig:vis_result_comparison} provides representative qualitative comparisons under the 10-context/20-target forecasting protocol. The examples show that the task is highly partially observed: even the ground-truth Sentinel-2 sequence contains frequent cloud-contaminated or missing frames, shown as white regions in the visualization. Despite this sparse supervision, the forecasts should still respond to the dense future meteorological forcing and predict the onset and progression of vegetation degradation.

The qualitative results are consistent with the quantitative trends in the experiments. Earthformer tends to produce conservative forecasts whose vegetation decline appears delayed relative to the observed target sequence, especially when heat and drought stress cause a rapid transition from green vegetation to dry or senescent surfaces. The diffusion-based models, Wan2.1 and EO-WM, better capture this forcing-response behavior and predict the degradation earlier. EO-WM further preserves more coherent spatial patterns and a more faithful timing of the response, which is consistent with the benefit of explicitly modeling climatological baseline, weather anomaly, and cumulative physical stress.

\subsection{Benchmark Construction and Evaluation Protocols}

\label{sec:supp_bench}


We describe the full construction pipelines for the two proposed benchmarks. Both are derived from existing EarthNet2021~\cite{requena2021earthnet2021} test splits via multi-stage quality filtering and event verification, producing curated evaluation sets that target specific capabilities beyond pixel-level fidelity.

\subsubsection{Extreme Summer Benchmark Construction}
\label{sec:supp_extreme}


\paragraph{Data source.}
The benchmark is constructed from the EarthNet2021 \texttt{extreme\_test} split, which targets the 2018 European heatwave and drought, one of the most severe compound climate events on record in Central Europe. This split contains 4{,}000 sequences of 60 frames (5-day cadence, $\sim$300 days) across four Sentinel-2 tiles (32UPC, 32UNC, 32UMC, 32UQC) covering parts of France and Germany. Each sequence provides four spectral bands (blue, green, red, NIR) at 20\,m resolution with an associated per-pixel quality mask indicating cloud and shadow contamination.

Our goal is to extract 30-frame evaluation windows (10 context $+$ 20 target frames) that each contain a \textbf{verified vegetation degradation event} in the target period, i.e., a transition from a stable, vegetated baseline to a significant Normalized Difference Vegetation Index (NDVI) decline driven by the extreme weather conditions.

\begin{figure*}[!h]
    \centering
    \includegraphics[width=0.99\textwidth]{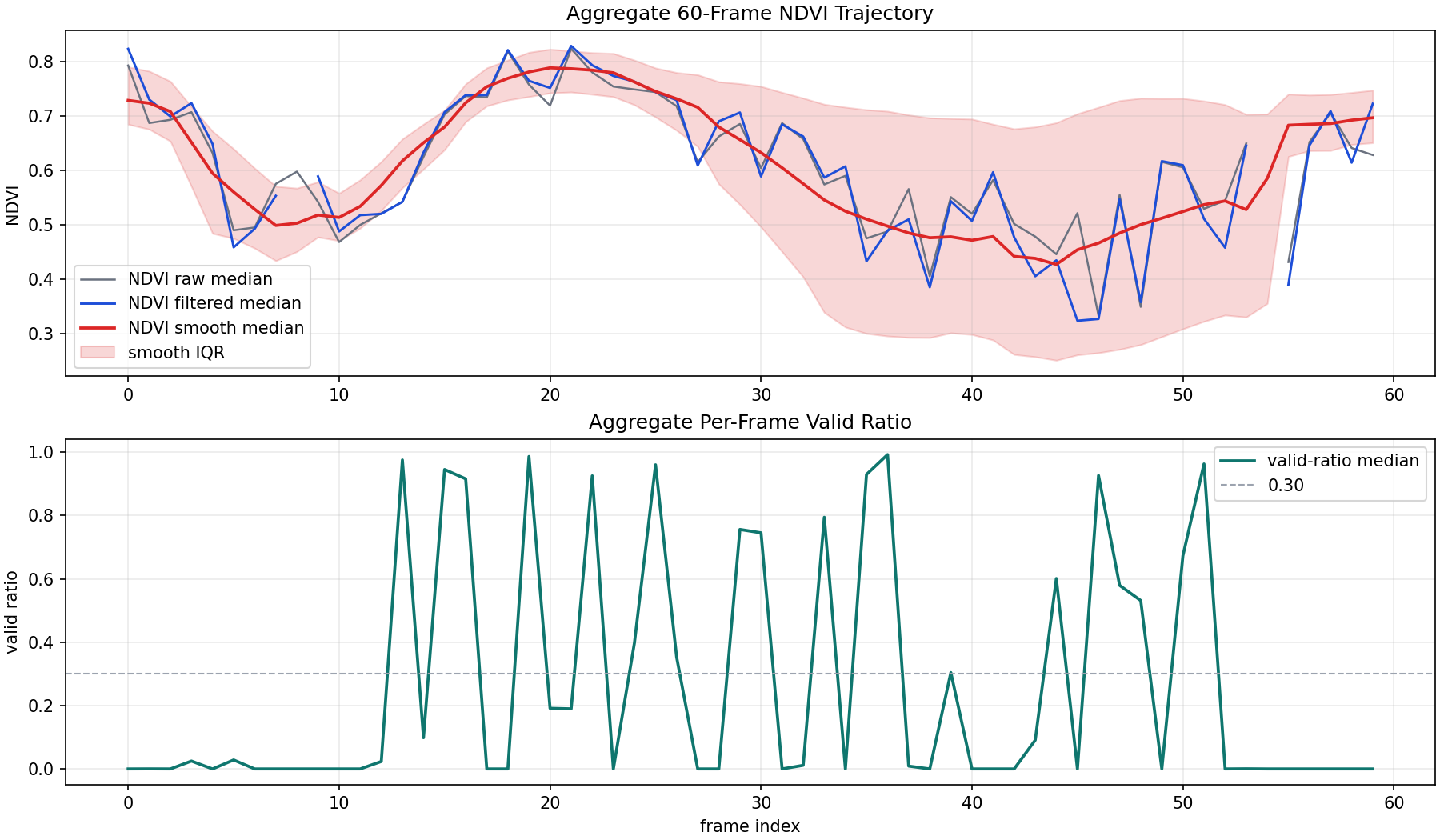}
    \caption{\textbf{Aggregate 60-frame diagnostics for Extreme Summer window construction.}
    Top: aggregate NDVI trajectory over the original 60-frame sequences. The gray curve shows the raw median NDVI computed from valid pixels, the blue curve shows the median after frame-level validity filtering and short-gap interpolation, and the red curve shows the smoothed median used for robust peak--trough detection. Bottom: aggregate per-frame valid-pixel ratio, with the dashed line marking the 0.30 validity threshold. The figure illustrates both the vegetation decline signal targeted by the benchmark and the sparse-observation regime caused by clouds and missing frames.}
    \label{fig:extreme_NDVI_aggregate_curve}
\end{figure*}

\paragraph{Stage 1: Sequence-level NDVI curve analysis.}
Guided by the aggregate behavior in Fig.~\ref{fig:extreme_NDVI_aggregate_curve}, we first analyze the full 60-frame NDVI trajectory of each sample to characterize its temporal dynamics and identify candidate sequences. For each frame, we compute the median NDVI over valid (cloud-free) pixels and apply a validity filter: frames with fewer than 30\% valid pixels are masked as unobservable. Short gaps ($\leq$3 consecutive masked frames) are filled by linear interpolation, while longer gaps are preserved as missing to avoid hallucinating trends. A 5-frame moving average is applied to obtain a smoothed trajectory for robust peak--trough detection.

From each trajectory, we extract statistics including the peak and trough NDVI values, the drop amplitude (peak minus trough), the number of local minima (with prominence $\geq 0.04$), the mean valid-pixel ratio in the context and target periods, and the temporal location of the global trough. 

\paragraph{Stage 2: Window localization and extreme-event verification.}

This stage identifies the optimal 30-frame window within each sequence and verifies that it contains a true extreme event. The procedure operates in two sub-stages.

\emph{Sub-stage 2a: Sample-level prefiltering.}
We apply three hard conditions to identify candidate sequences: (i) the mean valid-pixel ratio in the target period must exceed 30\%, ensuring sufficient observable data; (ii) the 60-frame sequence-level NDVI drop from context-end to target-minimum must be at least 0.35, indicating a substantial decline; and (iii) the global trough NDVI must be non-negative, excluding sensor artifacts. This prefilter is used only to select candidate sequences; the final-window drop amplitude $\Delta$ reported below is the peak-to-trough amplitude measured after window localization. Of 4{,}000 sequences, 1{,}447 pass all hard conditions and proceed to window-level analysis.

\emph{Sub-stage 2b: Anchor-based window placement.}
For each prefiltered sequence, we locate the transition point $t_{\text{stable}}$ where vegetation begins to decline, and place the 30-frame window so that $t_{\text{stable}}$ lies at or near the end of the context period. This is done via a \emph{global trough method}: we find the global NDVI minimum $t_{\text{trough}}$ and the preceding peak $t_{\text{peak}}$, then search backward from $t_{\text{trough}}$ for the latest frame where NDVI remains within 30\% of the peak value and local variability (standard deviation over a 5-frame neighborhood) is below 0.05. For sequences with complex multi-trough structure, we additionally apply a \emph{multi-trough method} that considers up to 8 local minima as candidate anchors and selects the one yielding the highest-confidence window.


\emph{Window quality filtering.}
Windows are discarded if they contain more than 6 invalid context frames (out of 10), more than 12 invalid target frames (out of 20), or more than 15 half-cloudy frames (valid ratio between 20\% and 80\%).

\emph{Extreme event verification.}
For each surviving window, we compute a baseline NDVI $\bar{y}_{\text{base}}$ as the mean NDVI of the last 4 valid context frames (requiring $\geq$20\% valid pixels per frame). We then set an extreme threshold $\theta = \bar{y}_{\text{base}} - 0.10$. A window is confirmed as containing an extreme event if at least 2 consecutive target frames and at least 2 total target frames have spatially-averaged NDVI below $\theta$ (each requiring $\geq$25\% valid pixels). Windows that fail this verification are rejected.

Each verified window receives a composite extreme score:
\begin{equation}
    s_{\text{extreme}} = \Delta \cdot (0.55 + 0.45\,r_{\text{persist}}) \cdot (0.70 + 0.30\,r_{\text{consec}}) \cdot (0.40 + 0.60\,q),
\end{equation}
where $\Delta$ is the NDVI drop amplitude, $r_{\text{persist}}$ is the fraction of valid target frames below threshold, $r_{\text{consec}} = \min(n_{\text{consec}}/6, 1)$ rewards sustained events, and $q$ is the mean valid-pixel ratio over the full window. The complete procedure is summarized in Algorithm~\ref{alg:extreme}.


\begin{algorithm}[h]
\caption{Extreme Summer Benchmark Construction}
\label{alg:extreme}
\begin{algorithmic}[1]
\Require EarthNet2021 \texttt{extreme\_test} split: 4{,}000 sequences $\times$ 60 frames
\Ensure Benchmark set $\mathcal{B}_{\text{ext}}$ of verified 30-frame extreme windows
\State \textbf{Stage 1: Curve analysis}
\For{each sequence $i = 1, \ldots, 4000$}
    \State Compute per-frame median NDVI; mask frames with $<$30\% valid pixels
    \State Interpolate NaN gaps $\leq$3 frames; smooth with 5-frame moving average
    \State Extract: peak/trough NDVI, drop amplitude, local minima count, valid ratios
\EndFor
\State \textbf{Stage 2a: Sample prefiltering}
\State Apply hard conditions: target valid ratio $\geq$0.30, sequence-level target drop $\geq$0.35, trough $\geq$0
\State $\mathcal{C} \leftarrow$ 1{,}447 sequences passing all conditions
\State \textbf{Stage 2b: Window localization \& verification}
\For{each sequence $i \in \mathcal{C}$}
    \State Locate anchor $t_{\text{stable}}$ via global-trough and/or multi-trough method
    \State Place 30-frame window with $t_{\text{stable}}$ near the context boundary
    \State Compute anchor confidence $c_{\text{anchor}}$
    \State \textbf{Filter:} reject if too many invalid/half-cloudy frames
    \State Compute baseline $\bar{y}_{\text{base}}$ from last 4 valid context frames
    \State \textbf{Verify:} confirm $\geq$2 consecutive and $\geq$2 total frames with NDVI $< \bar{y}_{\text{base}} - 0.10$
    \If{verified}
        \State Compute extreme score $s_{\text{extreme}}$; add to $\mathcal{B}_{\text{ext}}$
    \EndIf
\EndFor
\State \Return $\mathcal{B}_{\text{ext}}$ (1{,}440 windows), sorted by $s_{\text{extreme}}$
\end{algorithmic}
\end{algorithm}

\paragraph{Final benchmark statistics.}
Of 1{,}447 prefiltered candidates, 1{,}440 pass all verification criteria (7 rejected: 3 due to excessive context cloud cover, 3 failing extreme verification, 1 due to target cloud cover). Table~\ref{tab:extreme_stats} summarizes the benchmark statistics.

\begin{table}[h]
\centering
\caption{\textbf{Extreme Summer Benchmark statistics.} Distribution of key attributes across the 1{,}440 verified windows. Q1 denotes the 25th percentile.}
\label{tab:extreme_stats}
\small
\begin{tabular}{lcccccc}
\toprule
Attribute & Mean & Std & Min & Q1 & Median & Max \\
\midrule
Drop amplitude ($\Delta$) & 0.504 & 0.131 & 0.052 & 0.428 & 0.534 & 0.716 \\
Baseline NDVI ($\bar{y}_{\text{base}}$) & 0.743 & 0.055 & 0.434 & 0.713 & 0.746 & 0.890 \\
Target minimum NDVI & 0.287 & 0.091 & 0.150 & 0.213 & 0.260 & 0.676 \\
Extreme score ($s_{\text{extreme}}$) & 0.248 & 0.064 & 0.057 & 0.201 & 0.251 & 0.445 \\
Anchor confidence ($c_{\text{anchor}}$) & 0.898 & 0.062 & 0.544 & --- & --- & 0.999 \\
Valid-pixel ratio (mean) & 0.417 & 0.052 & 0.298 & 0.381 & 0.405 & 0.629 \\
\midrule
\multicolumn{7}{l}{\emph{Anchor mode distribution:} global-trough 62.0\%, multi-trough 27.2\%, exhaustive scan 10.7\%} \\
\multicolumn{7}{l}{\emph{Tile distribution:} 32UPC: 491, 32UNC: 416, 32UMC: 313, 32UQC: 220} \\
\bottomrule
\end{tabular}
\end{table}

\subsubsection{Seasonal Matched-Pair Benchmark Construction}
\label{sec:supp_seasonal}


\paragraph{Data source and design rationale.}
The Seasonal Matched-Pair Benchmark is constructed from the EarthNet2021 \texttt{seasonal\_test} split, which provides 3-year merged Sentinel-2 sequences (210 frames at 5-day cadence, covering 2017--2019) for each geographic location. By pairing windows from the \emph{same location} and \emph{same seasonal phase} but \emph{different years}, we isolate interannual weather variability as the sole driver of vegetation divergence: eliminating geographic and phenological confounds. This enables a direct ``what-if'' evaluation: given the same initial observation state, does the model produce appropriately different futures when driven by different weather forcing?

\paragraph{Window extraction.}
From each 210-frame sequence, we extract 30-frame windows (10 context $+$ 20 target frames, matching the model's operational setting). Each year spans frames 0--69, and we extract three windows per year at start offsets 0, 20, and 40 within the year (corresponding to early, mid, and late growing season phases), yielding up to 9 windows per location. For each window, we compute:
\begin{itemize}[nosep,leftmargin=12pt]
    \item Per-frame NDVI trajectory: median NDVI over valid pixels, with frames below 30\% validity masked, short gaps ($\leq$3 frames) interpolated, and a 5-frame moving average applied.
    \item Baseline NDVI: median of the last 4 input frames.
    \item Quality metrics: mean valid-pixel ratio for input and target periods, count of cloudy frames.
\end{itemize}
A window is marked as \emph{benchmark-usable} if both mean valid ratios exceed 20\% and the median valid pixel count exceeds 400 pixels per frame. From the full seasonal test split, 12{,}992 windows pass this soft quality gate.

\paragraph{Stage 1: Pair construction and divergence scoring.}

A pair $(A, B)$ is formed whenever two windows share the same geographic location (cube) and seasonal phase (window start offset within year) but come from different calendar years (e.g., 2017 vs.\ 2019). This yields 36{,}000 candidate pairs. For each pair, we compute multi-dimensional divergence scores capturing different aspects of how the two windows differ:

\emph{(a) Initial state distance ($D_{\text{init}}$).}
We measure how similar the two windows are at the start of the prediction period, combining: L2 distance of input NDVI trajectories (normalized by per-group standard deviation), NIR and Red trajectory distances, absolute difference of the last input NDVI value, spectral summary distance (mean band values), recent weather distance (last 3 input frames), and valid-ratio difference. These components are aggregated via equal-weighted summation after groupwise robust normalization (IQR-based standardization within each seasonal phase) to prevent cross-phase scale differences from dominating.

\emph{(b) Meteorological divergence ($D_{\text{meteo}}$).}
Captures how differently the weather evolves in the target period, combining: absolute differences in cumulative rainfall, mean/minimum/maximum temperature, peak daily maximum temperature, and a composite heat stress index. This score drives the \emph{Meteorological Divergence} track.

\emph{(c) Vegetation trajectory divergence ($D_{\text{veg}}$).}
Measures how differently the NDVI outcomes evolve, using a shift-tolerant ($\pm$3 frame) L2 trajectory distance that accounts for phenological timing differences, plus absolute differences in baseline-relative drop, recovery amplitude, and temporal volatility. This drives the \emph{Vegetation Trajectory Divergence} track.

\emph{(d) Pixel-level spatial divergence ($D_{\text{pixel}}$).}
The per-frame L1 distance of spatially co-registered NDVI values, averaged over frames with $\geq$30\% shared valid pixels (requiring $\geq$6 clean frames per pair). This drives the \emph{Pixel-level Spatial Divergence} track.

\paragraph{Stage 2: Quality filtering and initial-state matching.}

We apply a series of gates to ensure both data quality and experimental control:

\emph{Benchmark eligibility.} Both windows in a pair must independently pass the soft quality gate. This reduces the candidate pool from 36{,}000 to 8{,}444 pairs.

\emph{Initial-state matching.} To ensure that observed divergence in the target period reflects weather differences rather than different starting conditions, we retain only pairs whose $D_{\text{init}}$ falls at or below the 40th percentile within their seasonal phase group. This yields 3{,}379 closely-matched pairs that share similar initial vegetation states.

\emph{Hard quality gates.} We further require: (i) mean valid-pixel ratio $\geq$30\% for the full 30-frame window in both sides; (ii) $\leq$20 cloudy frames in each window; (iii) for the vegetation trajectory track, $\geq$8 clean frames with shared valid pixels; and (iv) for the pixel-level track, $\geq$6 clean frames. After these gates, the eligible pool sizes are: meteorological track 3{,}372, vegetation trajectory track 2{,}047, and pixel-level track 2{,}675.

\paragraph{Stage 3: Stratified selection and convergence.}

To produce a manageable and balanced benchmark, we apply stratified top-$N$ selection:
\begin{enumerate}[nosep,leftmargin=12pt]
    \item Within each track, we rank pairs by their respective divergence score (descending).
    \item Per seasonal phase (3 phases: offsets 0, 20, 40), we select the top 50 highest-divergence pairs, yielding 150 pairs per track.
    \item A per-cube cap of 3 pairs prevents any single location from dominating the benchmark.
    \item The three track selections are merged via union and deduplicated by pair identity. Each pair retains metadata indicating which track(s) selected it and its rank within each track.
\end{enumerate}

The final benchmark contains \textbf{422 unique pairs} (844 inference windows). Multi-track membership provides complementary evaluation perspectives: 394 pairs belong to a single paper-facing track, and 28 to two tracks. The complete procedure is summarized in Algorithm~\ref{alg:seasonal}, and Table~\ref{tab:seasonal_stats} summarizes the properties of the final 422-pair benchmark.

\begin{algorithm}[h]
\caption{Seasonal Matched-Pair Benchmark Construction}
\label{alg:seasonal}
\begin{algorithmic}[1]
\Require EarthNet2021 \texttt{seasonal\_test} split: 3-year sequences (210 frames), 5-day cadence
\Ensure Benchmark set $\mathcal{B}_{\text{sea}}$ of matched cross-year pairs with verified divergence
\State \textbf{Window extraction:} Extract 30-frame windows at 3 seasonal offsets $\times$ 3 years per location
\State Mark windows as usable if mean valid ratio $>$20\% and pixel count $>$400
\State \textbf{Pair construction:} Form all same-location, same-phase, cross-year pairs $\rightarrow$ 36{,}000 candidates
\For{each pair $(A, B)$}
    \State Compute $D_{\text{init}}$, $D_{\text{meteo}}$, $D_{\text{veg}}$, $D_{\text{pixel}}$ with groupwise normalization
\EndFor
\State \textbf{Stage 2: Filtering}
\State Require both windows benchmark-eligible $\rightarrow$ 8{,}444 pairs
\State Require $D_{\text{init}} \leq$ 40th percentile (per seasonal phase) $\rightarrow$ 3{,}379 pairs
\State Apply hard quality gates (valid ratio $\geq$0.30, cloudy $\leq$20, overlap frames) per track
\State \textbf{Stage 3: Stratified selection}
\For{each track $\in$ \{Meteorological, Vegetation Trajectory, Pixel-level\}}
    \For{each seasonal phase $\in$ \{0, 20, 40\}}
        \State Select top-50 pairs by track score, with per-cube cap of 3
    \EndFor
\EndFor
\State $\mathcal{B}_{\text{sea}} \leftarrow$ union of all track selections, deduplicated $\rightarrow$ 422 pairs
\State \Return $\mathcal{B}_{\text{sea}}$, with per-pair track membership and divergence metadata
\end{algorithmic}
\end{algorithm}

\begin{table}[t]
\centering
\caption{\textbf{Seasonal Matched-Pair Benchmark statistics.} (a)~Overall composition. (b)~Divergence score distributions across the selected pairs (scores are in normalized units after groupwise robust standardization).}
\label{tab:seasonal_stats}
\small
\begin{subtable}[t]{0.48\textwidth}
\centering
\caption{Composition}
\begin{tabular}{lr}
\toprule
Attribute & Value \\
\midrule
Total pairs & 422 \\
Unique geographic locations & 380 \\
Unique Sentinel-2 tiles & 11 \\
\midrule
\emph{Track membership} & \\
\quad Meteorological Divergence & 150 \\
\quad Vegetation Trajectory Divergence & 150 \\
\quad Pixel-level Spatial Divergence & 150 \\
\midrule
\emph{Seasonal phase distribution} & \\
\quad Phase 0 (early season) & 145 \\
\quad Phase 20 (mid season) & 142 \\
\quad Phase 40 (late season) & 135 \\
\bottomrule
\end{tabular}
\end{subtable}
\hfill
\begin{subtable}[t]{0.48\textwidth}
\centering
\caption{Divergence score distributions}
\begin{tabular}{lccc}
\toprule
Score & Mean & Median & Max \\
\midrule
$D_{\text{init}}$ & $-$0.505 & $-$0.466 & $-$0.220 \\
$D_{\text{meteo}}$ & 0.237 & 0.256 & 1.077 \\
$D_{\text{veg}}$ & 0.212 & 0.142 & 4.530 \\
$D_{\text{pixel}}$ & 0.478 & 0.259 & 3.951 \\
\bottomrule
\end{tabular}
\vspace{6pt}

{\small
$D_{\text{init}}$: negative values indicate close initial-state matching (lower = more similar).
$D_{\text{meteo}}$, $D_{\text{veg}}$, $D_{\text{pixel}}$: higher values indicate greater divergence.
All scores are in robust-normalized units (0 = group median, 1 $\approx$ 1 IQR above median).
}
\end{subtable}
\end{table}

\subsubsection{Extreme Summer Benchmark: Evaluation Metrics}
\label{sec:supp_extreme_metrics}


In addition to the standard EarthNetScore (ENS), we report reconstruction and event-severity metrics that are computed on the 20-frame target period of each benchmark window. When a method produces multiple stochastic samples, we follow the main-paper protocol and evaluate the ensemble-mean prediction. Deterministic methods have $N=1$.

\paragraph{Notation.}
Let $\hat{\mathbf{o}}^{(k)} \in \mathbb{R}^{C \times T \times H \times W}$, $k = 1, \ldots, N$, denote the $N$ generated target sequences and $\mathbf{o} \in \mathbb{R}^{C \times T \times H \times W}$ the ground truth, where $C$ is the number of spectral channels, $H$ and $W$ are the spatial dimensions, and $T=20$ is the target length. We use $c$ for channel index, $t$ for target-frame index, and $p$ for a spatial pixel index. The evaluated prediction is the ensemble mean $\bar{\hat{\mathbf{o}}}=N^{-1}\sum_{k=1}^{N}\hat{\mathbf{o}}^{(k)}$. Let $M_{t,p}\in\{0,1\}$ be the target validity mask, where 1 denotes an observable pixel. We write $y_{t,p}$ and $\hat{y}_{t,p}$ for NDVI computed from the ground truth and from $\bar{\hat{\mathbf{o}}}$, respectively. Let $\text{Red}_{t,p}$ and $\text{NIR}_{t,p}$ denote the ground-truth red and near-infrared channel values at frame $t$ and pixel $p$. Then:
\begin{equation}
    y_{t,p} = \frac{\text{NIR}_{t,p} - \text{Red}_{t,p}}{\text{NIR}_{t,p} + \text{Red}_{t,p} + \epsilon},
    \qquad \epsilon = 10^{-8}.
\end{equation}
The target-period vegetation mask is defined from the ground truth as
\begin{equation}
    \mathcal{V} = \{(t,p): y_{t,p} \geq 0.3,\; M_{t,p}=1\},
\end{equation}
and $\mathcal{V}_t=\{p:(t,p)\in\mathcal{V}\}$ denotes the valid vegetated pixels at frame $t$.

\paragraph{Pixel MAE (P-MAE).}
Pixel-MAE is the mean absolute error over all valid pixels and spectral channels:
\begin{equation}
    \text{P-MAE}
    =
    \frac{1}{C\sum_{t,p} M_{t,p}}
    \sum_{c=1}^{C}\sum_{t,p} M_{t,p}
    \left|\bar{\hat{o}}_{c,t,p}-o_{c,t,p}\right|.
\end{equation}

\paragraph{NDVI MAE.}
NDVI-MAE evaluates vegetation-state accuracy over valid vegetated pixels:
\begin{equation}
    \text{N-MAE} = \frac{1}{|\mathcal{V}|} \sum_{(t,p) \in \mathcal{V}} \left| \hat{y}_{t,p} - y_{t,p} \right|,
\end{equation}
where $\hat{y}_{t,p}$ is computed from $\bar{\hat{\mathbf{o}}}$.

\paragraph{Trough NDVI MAE (TN-MAE).}
TN-MAE is the NDVI error at the ground-truth trough frame $t^*$ recorded in the benchmark metadata:
\begin{equation}
    \text{TN-MAE} = \frac{1}{|\mathcal{V}_{t^*}|} \sum_{p \in \mathcal{V}_{t^*}} \left| \hat{y}_{t^*,p} - y_{t^*,p} \right|.
\end{equation}
This isolates prediction accuracy at the moment of maximum observed stress.

\paragraph{Drop Amplitude Error (DAE).}
For each window, the benchmark metadata provides a baseline NDVI $\bar{y}_{\text{base}}$ and a verified ground-truth drop amplitude $\Delta_{\text{gt}}$ from the construction stage (the stored \texttt{drop\_amplitude} field). The prediction-side target trajectory is computed as the frame-wise mean NDVI of the prediction over valid ground-truth vegetation pixels:
\begin{equation}
    \bar{\hat{y}}(t) = \frac{1}{|\mathcal{V}_t|}\sum_{p\in\mathcal{V}_t}\hat{y}_{t,p}.
\end{equation}
The predicted decline amplitude is
\begin{equation}
    \Delta_{\text{pred}} = \bar{y}_{\text{base}} - \min_{t:|\mathcal{V}_t|>0}\bar{\hat{y}}(t),
\end{equation}
and the Drop Amplitude Error is
\begin{equation}
    \text{DAE} = \left| \Delta_{\text{pred}} - \Delta_{\text{gt}} \right|.
\end{equation}
This measures how accurately the model reproduces the magnitude of the vegetation decline, regardless of exact timing.

\paragraph{Drop Reproduction Accuracy (DRA).}

A naive threshold-crossing detection metric can reward models that systematically under-predict NDVI. We therefore use DRA as an auxiliary severity-calibration score in the visualization analysis. Let $N_{\text{samples}}$ be the number of evaluated benchmark windows, and let $\Delta_{\text{pred}}^i$ and $\Delta_{\text{gt}}^i$ be the predicted and benchmark drop amplitudes for window $i$. Then:
\begin{equation}
    \text{DRA} = \frac{1}{N_{\text{samples}}} \sum_{i=1}^{N_{\text{samples}}} \max\!\left(0,\; 1 - \frac{|\Delta_{\text{pred}}^i - \Delta_{\text{gt}}^i|}{\Delta_{\text{gt}}^i}\right).
    \label{eq:dra}
\end{equation}
DRA$\,\in [0, 1]$ (higher is better). A score of 1.0 means the predicted decline amplitude exactly matches the benchmark drop amplitude, while predictions whose absolute error exceeds $\Delta_{\text{gt}}$ receive zero credit.

\paragraph{Severity-bin aggregation.}
The low-, mid-, and high-severity results in the main paper are obtained by splitting benchmark windows into three bins using the 33.3rd and 66.7th percentiles of the composite extreme score $s_{\text{extreme}}$, rather than by the drop amplitude alone. Within each bin, TN-MAE and DAE are averaged over per-window metric values.

\subsubsection{Seasonal Matched-Pair Benchmark: Evaluation Metrics}
\label{sec:supp_seasonal_metrics}


The Seasonal Matched-Pair metrics evaluate whether the model produces \emph{appropriately different} predictions when driven by different weather at the same location. Given a pair $(A, B)$ sharing the same geographic tile but observed in different years with divergent meteorological conditions, we compare the predicted vegetation trajectories against the ground-truth divergence. As above, stochastic methods are evaluated with the ensemble-mean prediction for each window. All metrics operate on the spatially-averaged NDVI trajectory over vegetation pixels ($\text{NDVI} \geq 0.3$ and valid).

\paragraph{Notation.}
For each window $w \in \{A, B\}$ in a pair, let $\bar{y}_{\text{gt}}^w(t)$ denote the ground-truth spatially-averaged NDVI at target frame $t$, and $\bar{\hat{y}}^w(t)$ the predicted NDVI from the ensemble-mean forecast, spatially averaged over vegetation pixels. Define the per-frame ground-truth divergence $d_t^{\text{gt}} = \bar{y}_{\text{gt}}^A(t) - \bar{y}_{\text{gt}}^B(t)$ and predicted divergence $d_t^{\text{pred}} = \bar{\hat{y}}^A(t) - \bar{\hat{y}}^B(t)$. Let $\mathcal{T}$ denote the set of target frames where both sides have valid observations (finite NDVI values). Here $t \in \{1,\dots,T_{\text{out}}\}$ indexes the target frames, $i$ indexes benchmark pairs, $\mathbf{1}\{\cdot\}$ denotes the indicator function, $\operatorname{sign}(\cdot)$ returns the sign of its argument, and $\rho_S(\cdot,\cdot)$ denotes Spearman's rank correlation.

\paragraph{Divergence Reproduction Ratio (DRR).}

DRR measures whether the model reproduces the correct \emph{magnitude} of vegetation divergence between paired windows. We compute absolute divergences and filter by a noise threshold $\tau = 0.02$ to exclude frames where the ground-truth difference is negligible:
\begin{equation}
    \text{DRR} = \frac{\overline{|d_t^{\text{pred}}|}_{\,t \in \mathcal{T}_\tau}}{\overline{|d_t^{\text{gt}}|}_{\,t \in \mathcal{T}_\tau}}, \quad \text{where } \mathcal{T}_\tau = \{t \in \mathcal{T} : |d_t^{\text{gt}}| > \tau\},
\end{equation}
and the overline denotes temporal averaging. Following the main-paper tables, we report DRR$_{\mathrm{mean}}$, the mean of the per-pair DRR values; the median DRR is also computed as a robust diagnostic. DRR $= 1.0$ is ideal: values below 1 indicate under-response (the model fails to differentiate sufficiently between different weather conditions), while values above 1 indicate over-response.

\paragraph{Directional Hit Rate (DHR).}

DHR measures whether the model correctly predicts \emph{which window has lower NDVI} at each timestep, conditioned on the ground-truth difference exceeding the noise floor:
\begin{equation}
    \text{DHR} = \frac{1}{|\mathcal{T}_\tau|} \sum_{t \in \mathcal{T}_\tau} \mathbf{1}\!\left\{\operatorname{sign}(d_t^{\text{pred}}) = \operatorname{sign}(d_t^{\text{gt}})\right\}.
\end{equation}
DHR $\in [0, 1]$ (higher is better). A score of 0.5 corresponds to random guessing; values significantly above 0.5 indicate that the model's response to different weather conditions is directionally correct. We aggregate DHR across all pairs by pooling the hit counts: $\text{DHR}_{\text{agg}} = \sum_i n_{\text{hits}}^i \,/\, \sum_i |\mathcal{T}_\tau^i|$, where $n_{\text{hits}}^i$ is the number of correct-sign timesteps for pair $i$ and $\mathcal{T}_\tau^i$ is its thresholded valid-frame set. This weights each pair proportionally to its number of valid comparison frames.

\paragraph{Paired Divergence Correlation (PDC).}

PDC evaluates whether pairs that exhibit large divergence in the real world also produce large divergence in the model's predictions (a \emph{ranking} fidelity measure across the full pair set). For each pair $i$, let $\mathcal{T}_i$ denote its valid comparison-frame set. We compute the total (time-summed) absolute divergence:
\begin{equation}
    D_i^{\text{gt}} = \sum_{t \in \mathcal{T}_i} |d_{i,t}^{\text{gt}}|, \qquad D_i^{\text{pred}} = \sum_{t \in \mathcal{T}_i} |d_{i,t}^{\text{pred}}|.
\end{equation}
PDC is the Spearman rank correlation between the two vectors across all $N_{\text{pairs}}$ evaluated pairs:
\begin{equation}
    \text{PDC} = \rho_S\!\left(\{D_i^{\text{gt}}\}_{i=1}^{N_{\text{pairs}}},\;\{D_i^{\text{pred}}\}_{i=1}^{N_{\text{pairs}}}\right).
\end{equation}
PDC $\in [-1, 1]$ (higher is better). A high PDC indicates that the model's sensitivity to weather variation is correctly calibrated across different magnitudes of forcing change: it responds more when the weather truly differs more, and less when the difference is modest.

\paragraph{Complementarity of the three metrics.}
The three metrics capture orthogonal aspects of weather-response fidelity:
\begin{itemize}[nosep,leftmargin=12pt]
    \item \textbf{DRR}: magnitude calibration (``how much'' divergence is reproduced);
    \item \textbf{DHR}: directional accuracy (``which way'' the divergence goes);
    \item \textbf{PDC}: ranking fidelity (``relative ordering'' of divergence across pairs).
\end{itemize}
A model could achieve high DHR (correct sign) while having poor DRR (under-responding in magnitude), or high PDC (correct ranking) while systematically under-predicting absolute divergence. Together, the three metrics provide a comprehensive assessment of whether the model functions as a faithful weather-conditioned world model that correctly translates exogenous forcing differences into appropriate surface-state differences.

\paragraph{Track-level evaluation.}
All metrics are computed both globally (across all 422 pairs) and per track (Meteorological Divergence, Vegetation Trajectory Divergence, Pixel-level Spatial Divergence). 

\subsubsection{Probabilistic Calibration Diagnostics}
\label{sec:supp_prob_calibration}

The main tables evaluate the five-sample ensemble mean because this gives a non-oracle point forecast from each stochastic model. To further assess the full predictive distribution, we compute sample-based uncertainty diagnostics on the same five generated forecasts. Let $z_i$ denote a ground-truth target NDVI value over valid vegetation pixels and let $\hat{z}_i^{(k)}$, $k=1,\ldots,K$, denote the $K=5$ sampled predictions. We estimate CRPS by
\begin{equation}
    \mathrm{CRPS}_i
    =
    \frac{1}{K}\sum_{k=1}^{K}\left|\hat{z}_i^{(k)}-z_i\right|
    -
    \frac{1}{2K^2}\sum_{k=1}^{K}\sum_{k'=1}^{K}
    \left|\hat{z}_i^{(k)}-\hat{z}_i^{(k')}\right|.
\end{equation}
We also report the spread-skill ratio, defined as the mean ensemble standard deviation divided by the RMSE of the ensemble mean, and the empirical 90\% quantile coverage. CRPS is better when lower; the spread-skill ratio and 90\% coverage are best when close to 1 and 0.9, respectively.

\begin{table}[h]
\centering
\caption{\textbf{Probabilistic calibration diagnostics for five-sample ensembles.} Metrics are computed on target-period NDVI over valid vegetation pixels. Lower CRPS is better. Spread-skill ratio and 90\% coverage are best when closer to 1 and 0.9, respectively.}
\label{tab:prob_calibration}
\small
\begin{tabular}{lccc}
\toprule
Method & CRPS$\downarrow$ & Spread-skill ratio $\rightarrow 1$ & 90\% coverage $\rightarrow 0.9$ \\
\midrule
Wan2.1-Fun-V1.1-1.3B-InP & 0.0300 & 0.5783 & 0.3816 \\
EO-WM (Ours) & \textbf{0.0249} & \textbf{0.6447} & \textbf{0.5210} \\
\bottomrule
\end{tabular}
\end{table}

Both five-sample ensembles remain under-dispersed, as indicated by spread-skill ratios below 1 and 90\% coverages below 0.9. However, EO-WM improves all three diagnostics relative to Wan2.1, reducing CRPS while increasing spread-skill ratio and empirical coverage. These results support the use of ensemble-mean evaluation in the main tables while making clear that the model is not perfectly calibrated.

\subsection{Comparison Method Adaptations}

\label{sec:supp_comparison}

\subsubsection{Wan2.1 Adaptation}
\label{sec:supp_wan}

To evaluate whether a strong general-purpose video generation model can serve as an effective baseline for Earth surface forecasting, we adapt Wan2.1-Fun-V1.1-1.3B-InP~\cite{wan2025wan} (abbreviated as Wan2.1-Inp), a 1.3B-parameter latent video diffusion transformer originally designed for video inpainting and prediction, to the EarthNet2021 task. This adaptation requires non-trivial architectural modifications to handle 4-channel multi-spectral input (B, G, R, NIR) and to inject geospatial and meteorological conditioning, followed by a carefully staged fine-tuning procedure to preserve the pretrained generative capabilities while specializing to the satellite domain.

\paragraph{Base model architecture.}
Wan2.1-Inp is built on a 3D Diffusion Transformer (DiT) with flow matching training~\cite{lipman2022flow}. The architecture consists of 30 transformer blocks with hidden dimension 1536, 12 attention heads, and a feed-forward dimension of 8960. Video inputs are patchified with a 3D patch size of $(1, 2, 2)$ (temporal, height, width). The model uses a causal 3D VAE that compresses video spatially by $8{\times}$ and temporally by $4{\times}$, producing 16-channel latent representations. As an inpainting model, it accepts both noisy latents and condition-frame latents concatenated along the channel dimension (input dimension $= 16 + 16 + 1_{\text{mask}} + 3_{\text{pad}} = 36$). Conditioning is injected via AdaLN modulation from the diffusion timestep embedding, and text guidance is provided through cross-attention from a T5-based text encoder.

\paragraph{Architectural modifications for EO adaptation.}
We introduce two categories of modifications to adapt the model for the EarthNet2021 satellite prediction task:

\textit{(1) VAE channel expansion.}
The original VAE operates on 3-channel RGB video. We extend it to 4-channel input (B, G, R, NIR) by expanding the encoder's first convolutional layer from $\text{CausalConv3d}(3 \to d)$ to $\text{CausalConv3d}(4 \to d)$, and the decoder's output layer from $\text{CausalConv3d}(d \to 3)$ to $\text{CausalConv3d}(d \to 4)$. The weights for the new 4th channel (NIR) are initialized as the average of the Red and Green channel weights, providing a reasonable starting point that leverages the pretrained spectral representations.

\textit{(2) Earth observation conditioning modules.}
We add four conditioning pathways to the DiT to inject geospatial and meteorological information:
\begin{itemize}[nosep,leftmargin=12pt]
    \item \emph{Global geospatial embedding.} A 6-dimensional vector encoding the tile's geographic location (3D spherical coordinates from latitude/longitude) and temporal position (cyclical day-of-year encoding and normalized year) is projected through a 2-layer MLP and added to the diffusion timestep embedding before AdaLN modulation, providing location- and season-aware generation.
    \item \emph{Per-frame temporal embedding.} A 3-dimensional per-frame vector (cyclical day-of-year, normalized year) is projected to the hidden dimension and added to each token according to its temporal position, enabling frame-level temporal awareness.
    \item \emph{DEM spatial encoder.} A 3-layer stride-2 convolutional encoder processes the digital elevation model (1 channel + validity mask) from $128{\times}128$ to $16{\times}16$ feature maps, which are broadcast across the temporal dimension, patchified to match the transformer's token grid, linearly projected, and added to the token embeddings.
    \item \emph{ERA5 meteorological encoder.} A 3-layer stride-2 convolutional encoder processes 5-channel ERA5 reanalysis data (precipitation, pressure, mean/min/max temperature, plus a validity mask) per frame. The resulting spatiotemporal features are patchified, projected, and added to the token stream alongside the DEM features.
\end{itemize}
All new modules are initialized with near-zero output (zero bias, small-norm weights) to ensure that the pretrained generation behavior is preserved at initialization.

\paragraph{Four-stage fine-tuning procedure.}
To transfer the strong video generation prior of Wan2.1 to the satellite domain without catastrophic forgetting, we design a progressive four-stage fine-tuning strategy that gradually unfreezes model capacity:

\textit{Stage~1: VAE warm-up.}
Only the newly expanded input and output convolutional layers of the VAE are trained (approximately 2M parameters), while all intermediate encoder and decoder layers remain frozen. This allows the NIR channel weights to adapt to the VAE's internal feature space without disrupting the pretrained reconstruction capability for the RGB channels. The model is trained for 5 epochs with learning rate $5 \times 10^{-5}$ using MSE reconstruction loss with KL regularization (weight $10^{-6}$), masked by per-pixel quality flags to exclude cloud-contaminated observations.

\textit{Stage~2: VAE full fine-tuning.}
Starting from the Stage~1 checkpoint, all VAE parameters (approximately 100M) are unfrozen and trained for 30 epochs at a reduced learning rate of $1 \times 10^{-5}$ with cosine annealing. This allows the intermediate layers to fully adapt to the statistical properties of 4-channel satellite imagery. After this stage, we recompute the per-channel latent normalization statistics (mean and standard deviation across 2{,}000 training samples) to replace the original RGB-video statistics, ensuring stable DiT training.

\textit{Stage~3: DiT LoRA warm-up.}
The pretrained DiT weights are frozen, and we train only the newly added EO conditioning modules (approximately 2M parameters) together with rank-32 LoRA adapters~\cite{hulora} attached to the query, key, value, and feed-forward projections of all transformer blocks (approximately 15M parameters). This stage runs for 20{,}000 steps with learning rate $1 \times 10^{-4}$, effective batch size 32, and cosine scheduling with 500 warm-up steps. The quality mask is applied to the diffusion loss to exclude invalid pixels.  

\textit{Stage~4: DiT full fine-tuning.}
Starting from the merged LoRA checkpoint, all 1.3B DiT parameters are unfrozen and trained for 50{,}000 steps at a lower learning rate of $5 \times 10^{-6}$ to refine the full model. An exponential moving average (EMA) with decay 0.9999 is maintained for stable evaluation. Gradient checkpointing and bfloat16 mixed precision are used throughout to fit within GPU memory.

\paragraph{Training hyperparameters.}
Table~\ref{tab:wan_hyper} summarizes the key hyperparameters across all four training stages.

\begin{table}[ht!]
\caption{\textbf{Wan2.1-Inp fine-tuning hyperparameters} across four progressive stages.}
\centering
\label{tab:wan_hyper}
\small
\begin{tabular}{@{}lcccc@{}}
\toprule
                         & Stage 1       & Stage 2       & Stage 3         & Stage 4         \\
                         & (VAE warm-up) & (VAE full)    & (DiT LoRA)      & (DiT full)      \\ \midrule
Trainable params         & 2M            & 100M          & 17M             & 1.3B            \\
Optimizer                & \multicolumn{4}{c}{AdamW}    \\
Learning rate            & $5{\times}10^{-5}$ & $1{\times}10^{-5}$ & $1{\times}10^{-4}$ & $5{\times}10^{-6}$ \\
Effective batch size     & 16            & 16            & 32              & 32              \\
Training duration        & 5 epochs      & 30 epochs     & 20k steps       & 50k steps       \\
LR scheduler             & \multicolumn{4}{c}{Cosine annealing}                             \\
Warmup steps             & ---           & ---           & 500             & 500             \\
Gradient norm clip       & \multicolumn{4}{c}{1.0}                                          \\
Mixed precision          & \multicolumn{4}{c}{bfloat16}                                     \\
EMA decay                & ---           & ---           & ---             & 0.9999          \\
LoRA rank                & ---           & ---           & 32              & ---             \\
Gradient checkpointing   & ---           & ---           & \checkmark      & \checkmark      \\
\bottomrule
\end{tabular}
\end{table}

\paragraph{Inference configuration.}
At inference time, we use 30-step Euler ODE integration with a flow-matching scheduler (shift${}=5.0$). The model receives 10 context frames and generates 20 future frames at $128{\times}128$ resolution with 4 spectral channels. Classifier-free guidance is disabled for all Wan2.1 evaluations (guidance scale${}=1.0$).

\subsubsection{Latte Adaptation}

To make Latte~\cite{ma2024latte} condition-consistent with the EO forecasting setting, we adapt it with an explicit cross-attention pathway for non-visual EO conditions. We keep Latte's video diffusion transformer backbone and repaint-style video-to-video prediction, but augment each transformer block with a lightweight cross-attention module whose keys and values are produced from future meteorological and geospatial condition tokens. This turns Latte from a visual-context-only video diffusion baseline into a weather-conditioned latent video diffusion model.

The adaptation follows a two-stage training procedure. First, we train a 4-channel VAE from scratch on EarthNet2021 B/G/R/NIR frames, using masked reconstruction loss over valid pixels. The VAE uses 4 latent channels and is trained with learning rate $10^{-4}$ and KL weight $10^{-6}$ until convergence, which occurs at approximately 50k steps. Second, we train a Latte-XL/2 DiT in the learned latent space for 30-frame EarthNet clips, matching the 10-context/20-target protocol used by the other methods. The DiT uses the Latte-XL/2 configuration with patch size 2, 28 transformer blocks, hidden size 1152, 16 attention heads, learned variance prediction, and 250 DDIM sampling steps at evaluation.

For condition injection, ERA5 variables are sampled at the Sentinel-2 frame times and encoded by a small spatiotemporal convolutional encoder at the latent resolution. DEM is encoded by a separate spatial convolutional encoder and broadcast across time. Geospatial and seasonal metadata are projected by an MLP and appended as global condition tokens. The resulting condition sequence is projected to the Latte hidden dimension and used as the encoder context for cross-attention after the self-attention sublayer in each spatial transformer block. The cross-attention output is zero-initialized through a gated residual projection.

At inference time, the 10 observed context frames are encoded into latents and reinserted during the reverse diffusion process, while the remaining 20 latent frames are sampled under the same future ERA5, DEM, and metadata conditions used by the other EO forecasting baselines. We train the conditioned Latte variant for 200k steps with AdamW, learning rate $10^{-4}$, local batch size 4, and 1k warm-up steps.

\subsubsection{OpenSTL Deterministic Baselines}
\label{sec:supp_openstl}

We evaluate six deterministic spatiotemporal prediction baselines from OpenSTL~\cite{tan2023openstl}: SimVP~\cite{gao2022simvp}, TAU~\cite{tan2023temporal}, PredRNN~\cite{wang2017predrnn}, PredRNNv2~\cite{wang2022predrnn}, and PhyDNet~\cite{guen2020disentangling}. These methods cover CNN-based, recurrent, and physics-informed prediction paradigms and are adapted to the EarthNet2021 setting with 10 observed frames and 20 future frames at $128{\times}128$ resolution.

\paragraph{EarthNet input/output adaptation.}
Each input frame contains 10 channels: 4 Sentinel-2 satellite bands (B, G, R, NIR), 5 ERA5 mesodynamic variables, and a static elevation channel. The ERA5 variables are temporally aligned to the Sentinel-2 timestamps and spatially upsampled to $128{\times}128$; the static elevation channels are repeated over time. All models are trained to predict the 4 satellite bands. For training, the target tensor additionally carries a cloud/quality mask and the 6-channel condition sequence.

\paragraph{FiLM conditioning.}
In the film OpenSTL adaptation, meteorological and static conditions are explicitly provided during the prediction horizon, instead of conditioning only on the observed-context weather. For SimVP, the encoder and decoder operate on the 4 satellite channels, while the 6 auxiliary channels are encoded by a small convolutional condition encoder and injected into the temporal processor through FiLM~\cite{perez2018film} modulation. For recurrent models, including PredRNN and PredRNNv2, we apply per-frame FiLM before the recurrent backbone. The full 30-frame condition sequence is formed by concatenating the 10 observed-context conditions with the 20 future conditions, and each satellite frame is modulated with the condition at the corresponding time step. PhyDNet uses the same per-frame FiLM mechanism for both context frames and teacher-forced target frames during training; at inference, the future FiLM modulation still injects weather information through the generated sequence. TAU follows the OpenSTL SimVP-style implementation and receives the 6 condition channels by concatenation at each rollout.

\paragraph{Prediction rollout.}
For SimVP and TAU, which naturally predict an output sequence with the same length as the input sequence, the 20-frame horizon is generated by two 10-frame auto-regressive rollouts. The first rollout predicts frames 11--20 from the observed frames 1--10 using future conditions for frames 11--20. The second rollout predicts frames 21--30 from the previously predicted satellite frames, with the future conditions for frames 21--30 reattached. Recurrent and physics-informed methods generate the full 20-frame horizon using their native OpenSTL training and inference procedures, with the same future-condition sequence supplied through FiLM.

\paragraph{Training details.}
All OpenSTL baselines are trained from scratch on the EarthNet2021 training set for 200 epochs. We use cloud-masked MSE loss on the 4 satellite output channels, where pixels marked invalid by the quality mask are excluded from the effective loss. The film training recipe uses AdamW, weight decay $10^{-5}$, cosine learning-rate scheduling, 10 warm-up epochs, minimum learning rate $10^{-5}$, and gradient clipping at 1.0. The best checkpoint is selected by validation loss and used for all evaluations. 


\subsection{Data and Asset Availability}
\label{sec:supp_assets}

\paragraph{Released benchmark files.}
The proposed Extreme Summer and Seasonal Matched-Pair benchmarks are derived from the public EarthNet2021 test splits. At submission time, we provide the inference CSV files used in our experiments, which specify the selected benchmark windows, pair identities, split membership, track labels, and metadata needed to run evaluation on the corresponding EarthNet2021 samples. The CSV files do not redistribute the raw EarthNet2021 imagery or weather data, users should obtain the raw data from the official EarthNet2021 source and apply the terms of the original dataset.


\paragraph{Existing assets and licenses.}
We use and cite publicly available datasets, model weights, and codebases. EarthNet2021~\cite{requena2021earthnet2021} is distributed under the CC-BY-NC-SA 4.0 license according to its official dataset page (\url{https://earthnet.tech/resources/datasets/earthnet2021}). EarthNet2021 includes Copernicus Sentinel data, whose access and use are governed by the Copernicus free, full, and open data policy as documented by the EarthNet2021 dataset page. The EO-VAE tokenizer~\cite{lehmann2026eo} is listed as Apache-2.0 on its official Hugging Face release (\url{https://huggingface.co/nilsleh/eo-vae}). Open-Sora~\cite{zheng2025open} is released under Apache-2.0 on its official GitHub repository (\url{https://github.com/hpcaitech/Open-Sora}). Wan2.1-Fun-V1.1-1.3B-InP~\cite{wan2025wan} is listed as Apache-2.0 on the official Alibaba-PAI Hugging Face release (\url{https://huggingface.co/alibaba-pai/Wan2.1-Fun-V1.1-1.3B-InP}). Latte~\cite{ma2024latte}, OpenSTL~\cite{tan2023openstl}, and Earthformer~\cite{gao2022earthformer} are also released under Apache-2.0 according to their official repositories (\url{https://github.com/Vchitect/Latte}, \url{https://github.com/chengtan9907/OpenSTL}, and \url{https://github.com/amazon-science/earth-forecasting-transformer}). We use these assets only for research benchmarking and adaptation, and we respect the corresponding attribution and license terms.

\end{document}